\documentclass{article}

\PassOptionsToPackage{numbers, compress}{natbib}

\usepackage[preprint]{neurips_2024}

\usepackage{amsmath,amsfonts,bm}

\def\eqref#1{equation~\ref{#1}}

\def\ceil#1{\lceil #1 \rceil}
\def\floor#1{\lfloor #1 \rfloor}
\def\1{\bm{1}}

\DeclareMathAlphabet{\mathsfit}{\encodingdefault}{\sfdefault}{m}{sl}
\SetMathAlphabet{\mathsfit}{bold}{\encodingdefault}{\sfdefault}{bx}{n}

\usepackage[utf8]{inputenc} 
\usepackage[T1]{fontenc}    
\usepackage{hyperref}       
\usepackage{url}            
\usepackage{booktabs}       
\usepackage{amsfonts}       
\usepackage{nicefrac}       
\usepackage{microtype}      
\usepackage{xcolor}         
\usepackage{graphicx}
\usepackage{hyperref}

\usepackage{url}
\usepackage{inconsolata}
\usepackage{highlight}
\usepackage{xcolor}
\usepackage{colortbl}
\usepackage{caption}
\usepackage{pgf}
\usepackage{textcomp}
\usepackage{graphicx}
\usepackage{subcaption}
\usepackage{mdframed}
\usepackage{booktabs}
\usepackage{cleveref}
\usepackage{subcaption}
\usepackage{multirow}
\usepackage{arydshln}
\graphicspath{{./figures/}}
\usepackage[breakable, theorems, skins]{tcolorbox}
\usepackage{seqsplit}
\usepackage{tikz}
\usepackage{wrapfig}

\DeclareRobustCommand{\StartCrate}{%
  \begingroup\normalfont
  \raisebox{-0.3ex}{\smash{\includegraphics[height=1.6\fontcharht\font`\B]{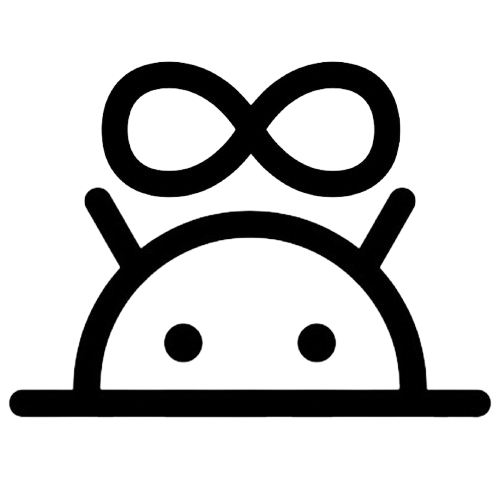}}}%
  \endgroup
}

\usepackage{algorithmicx}
\usepackage{algpseudocode}

\usepackage{algorithm}

\usepackage{listings}

\definecolor{commentcolor}{RGB}{110,154,155}   
\lstdefinestyle{mystyle}{
    backgroundcolor=\color{white},   
    commentstyle=\color{commentcolor},
    keywordstyle=\color{purple},
    numberstyle=\tiny\color{gray},
    stringstyle=\color{blue},
    basicstyle=\ttfamily\footnotesize,
    breakatwhitespace=false,         
    breaklines=true,
    columns=fullflexible,
    escapeinside=``,
    captionpos=b,                    
    keepspaces=true,                 
    numbers=left,                    
    numbersep=5pt,                  
    showspaces=false,                
    showstringspaces=false,
    showtabs=false,                  
    tabsize=4
}
\renewcommand{\mathbf}{\boldsymbol}

\makeatletter
\def\Ddots{\mathinner{\mkern1mu\raise\p@
\vbox{\kern7\p@\hbox{.}}\mkern2mu
\raise4\p@\hbox{.}\mkern2mu\raise7\p@\hbox{.}\mkern1mu}}
\makeatother

\newcommand{\norm}[2]{\left\| #1 \right\|_{#2}}
\newcommand{\abs}[1]{\left| #1 \right|}

\numberwithin{equation}{section}

\definecolor{citecolor}{HTML}{0071bc}
\hypersetup{
    colorlinks=true,%
    citecolor=citecolor,%
    filecolor=citecolor,%
    linkcolor=citecolor,%
    urlcolor=citecolor
}

\usepackage{enumitem}

\definecolor{darkgreen}{rgb}{0.0, 0.5, 0.0}

\input{macros.def}

\author{Hao Bai$^{1,2}$\thanks{Equal contribution, listed in alphabetical order; work done at UC Berkeley. E-mails: haob2@illinois.edu, yifei\_zhou@berkeley.edu, aviralkumar@google.com. Project page: \url{https://digirl-agent.github.io/}. Code available at \url{https://github.com/DigiRL-agent/digirl}.}  
\: Yifei Zhou$^{1*}$ 
\: Mert Cemri$^{1}$ 
\: Jiayi Pan$^{1}$ 
\AND 
\: Alane Suhr$^{1}$ 
\: Sergey Levine$^{1}$ 
\: Aviral Kumar$^{3}$ 
\AND \normalfont 
$^{1}$UC Berkeley
\: $^{2}$UIUC
\: $^{3}$Google DeepMind}

\title{\StartCrate{} {\ouragentnospace}: Training In-The-Wild Device-Control Agents with Autonomous Reinforcement Learning}

\begin{document}

\maketitle

\begin{abstract}
Training corpuses for vision language models (VLMs) typically lack sufficient amounts of decision-centric data.
This renders off-the-shelf VLMs sub-optimal for decision-making tasks such as in-the-wild device control through graphical user interfaces (GUIs). While training with static demonstrations has shown some promise, we show that such methods fall short for controlling real GUIs due to their failure to deal with real world stochasticity and non-stationarity not captured in static observational data. This paper introduces a novel autonomous RL approach, called \ouragentnospace, for training in-the-wild device control agents through fine-tuning a pre-trained VLM in two stages: offline RL to initialize the model, followed by offline-to-online RL. To do this, we build a scalable and parallelizable Android learning environment equipped with a VLM-based evaluator and develop a simple yet effective RL approach for learning in this domain. Our approach runs advantage-weighted RL with advantage estimators enhanced to account for stochasticity along with an automatic curriculum for deriving maximal learning signal. We demonstrate the effectiveness of {\ouragentnospace} using the Android-in-the-Wild (AitW) dataset, where our 1.3B VLM trained with RL achieves a 49.5\% absolute improvement -- from 17.7 to 67.2\% success rate -- over supervised fine-tuning with static human demonstration data. These results significantly surpass not only the prior best agents, including AppAgent with GPT-4V (8.3\% success rate) and the 17B CogAgent trained with AitW data (38.5\%), but also the prior best autonomous RL approach based on filtered behavior cloning (57.8\%), thereby establishing a new state-of-the-art for digital agents for in-the-wild device control.

\end{abstract}

\vspace{-0.2cm}
\section{Introduction}
\vspace{-0.2cm}


Advances in vision-language models (VLMs), especially in regards to their remarkable common-sense, reasoning, and generalization abilities imply that realizing a fully autonomous digital AI assistant, that can simplify human life by automating day-to-day activities on computer devices via natural language interfaces, is no longer a distant aspiration~\citep{koh2024visualwebarena, yan2023gpt4v, Zhou2023WebArenaAR}. An effective device-control AI assistant should be able to complete tasks in-the-wild through Graphical User Interfaces (GUIs) on digital devices: make travel plans; experiment with presentation designs; and operate a mobile device autonomously, all while running amidst stochasticity and distractors on the device, the Internet, and the tools it interacts with. However, enhanced reasoning or common-sense abilities do not directly transfer to intelligent assistant behavior: ultimately we want AI assistants to accomplish tasks, exhibit rational behavior, and recover from their mistakes as opposed to simply producing a plausible completion to a given observation based on the data seen during pre-training. This implies that a mechanism to channel abilities from pre-training into a deployable AI ``agent'' is lacking.

\begin{figure}[t]
     \centering
    \includegraphics[width=0.98\textwidth]{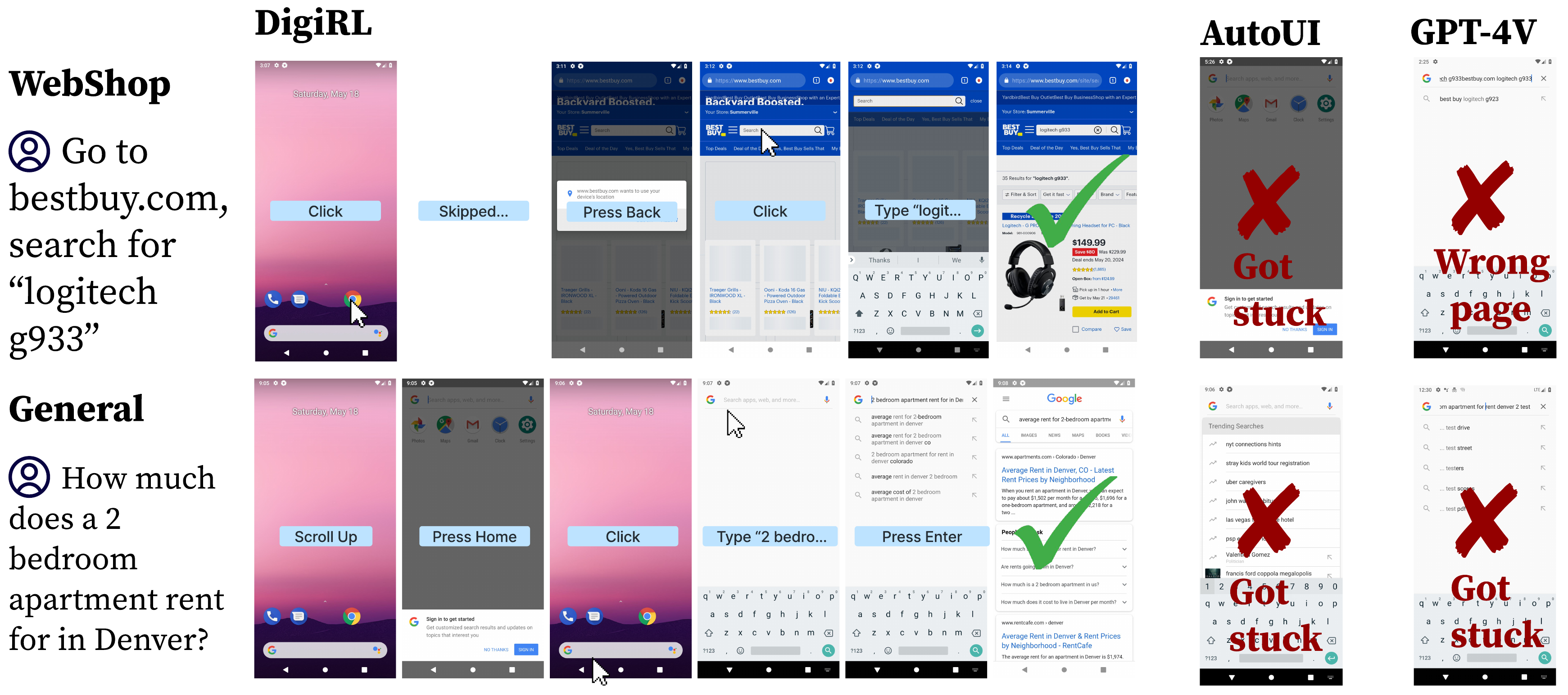}
    \caption{\footnotesize{\textbf{Qualitative comparison between {\ouragentnospace} and other approaches.} AutoUI trained from static human demonstrations can easily get stuck in out-of-distribution states while GPT-4V often get on a wrong goal (searched ``logitech g933bestbuy.com logitech g933'' in Google instead of bestbuy.com). In contrast, \ouragent can recover from such states and complete complex instruction as requested.}}
    \label{fig:example-compare}
    \vspace{-0.2cm}
\end{figure}

Even the strongest proprietary VLMs, such as GPT-4V~\citep{gpt4} and Gemini 1.5 Pro~\citep{geminiteam2024gemini1.5}~\footnote{We use external versions of these models as of June 11, 2024. Experiments with GPT and Gemini models were performed entirely by Hao Bai, Yifei Zhou, Mert Cemri, and Jiayi Pan.}, still struggle to produce the right actions when completing tasks on devices. While general-purpose vision-language abilities help these models still make meaningful abstract deductions about novel scenes when deployed, these deductions do not transfer to accurate reasoning for control~\citep{yang2023appagent, yan2023gpt4v, zheng2024gpt4vision, xie2024osworld}. As a result, most prior work for building device agents construct complex wrappers around proprietary VLMs by combining them with prompting, search, or tool use~\citep{yang2023appagent, xie2024osworld, zhang2024android, zhang2024ufo, yan2023gpt4v}. While building prompting or retrieval wrappers to improve decision-making performance of existing VLMs enhances their performance in the short run, without updating the weights, the effectiveness of the resulting agent is inherently limited by the capabilities of the base model~\cite{agenttuning, Chen2023FireActTL}. For example, we found that off-the-shelf VLMs make reasoning failures that derail the agent (e.g.,  Figure~\ref{fig:example-compare} and Figure~\ref{fig:gpt4_search_failure}), as direct consequences of inability of the base model to reason with low-level device-control actions. 
A different solution is to fine-tune the model on demonstrations via imitation learning. However, the dynamic nature of the web and device means that models trained to mimic actions in stale data can result in sub-optimalilty as the eco-system changes~\citep{pan2024autonomous}. Agents trained in this way struggle to recover from the agents' own mistakes~\citep{ghosh2021generalization,jiang2024importance}.

\begin{figure}[t]
     \centering
    \includegraphics[width=0.90\textwidth]{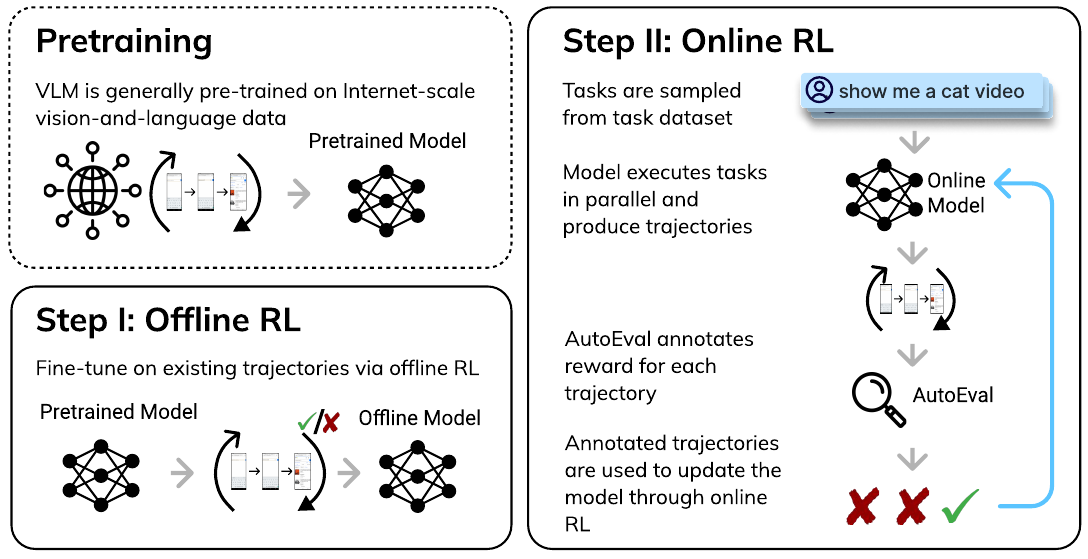}
    \caption{\footnotesize{{\bf {\ouragentnospace} overview.} {\ouragentnospace} is built upon a VLM that has been pre-trained on extensive web data to develop fundamental skills such as common knowledge, reasoning, and visual grounding. Initially, we employ offline RL to fine-tune the VLM using stale task-specific data, which helps in eliciting goal-oriented behaviors. Subsequently, our agent engages with real-world graphical user interfaces, continuously enhancing its performance through online RL and autonomous performance evaluations.}}
    \label{fig:teaser}
\end{figure}

If we can instead build an interactive approach to \emph{train} a VLM to directly adapt and learn \emph{from its own experience} on the device and the Internet, that can be used to build a robust and reliable device-control agent, without needing wrappers on top of proprietary models. However, this learning-based approach must satisfy some desiderata. First, it must make use of online interaction data since static demonstration data would not be representative of the task when the model is deployed: for instance, even in the setting of web navigation alone, dynamic nature of in-the-wild websites means that the agent will frequently encounter website versions that differ significantly from the scenarios seen during training and will need to behave reliably despite changes in visual appearance and distractions. Second, learning on-the-fly means the approach must learn from multi-turn interaction data from the model itself, a large of chunk of which would consist of failures. Proper mechanisms must be designed to automatically pick out the correct actions while filtering the wrong ones.

To this end, \textbf{our main contribution} is a novel autonomous RL approach, \ouragent (i.e., RL for Digital Agents), for training device control agents, as shown in~\Cref{fig:teaser}. The resulting agent attains state-of-the-art performance on a number of Android device-control tasks. To train this agent, our approach operates in two phases: an initial offline RL phase to initialize the agent using existing data, followed by an offline-to-online RL phase, that further fine-tunes the model obtained from offline RL on online rollout data. Online RL training requires access to an environment that the agent can interact with and obtain reliable reward signals, all in a reasonable amount of wall-clock time. To do so, we build a scalable and parallelizable Android learning environment equipped with a robust VLM-based general-purpose evaluator~\citep{pan2024autonomous} (average error rate 2.8\% against human judgement) that supports running up to 64 real Android emulators at the same time to make online RL real-time. Then, to effectively learn autonomously, we develop an online RL approach that retains the simplicity of supervised learning, but incorporates several key deep RL insights to enable fast fine-tuning. Concretely, our approach is a variant of advantage-weighted regression (AWR)~\citep{awr}, equipped with: \textbf{(i)} an automatic curriculum that uses an instruction-level value function to order tasks so as to extract maximal learning signal, which is inspired by prioritized replay methods~\citep{plr, schaul2016prioritized, openai2019solving}, and \textbf{(ii)} another step-level value function trained via effective cross-entropy loss~\citep{kumar2023offline, farebrother2024stop} to extract low-variance and less-biased learning signal amidst stochasticity and diverse tasks. This RL approach allows us to fine-tune VLMs on their own experience.

We evaluate our agent trained with \ouragent in carrying out diverse instructions from \textbf{Android in the Wild dataset}~\citep{aitw} on real Android device emulators and find that our agent can achieve a \textbf{28.7\% improvement} over the existing state-of-the-art agents (from 38.5\% to 67.2\% success rate) 18B CogAgent~\citep{hong2023cogagent}, and over 9\% improvement over the prior best autonomous learning approach based on Filtered Behavior Cloning~\citep{lai2024autowebglm, pan2024autonomous}. 
The performance of our agent also significantly surpasses wrappers on top of state-of-the-art proprietary VLMs such as GPT-4V~\citep{gpt4} and Gemini 1.5 Pro~\citep{geminiteam2024gemini1.5} (17.7\% success rate),
despite using a significantly smaller model (with 1.3B parameters). To our knowledge, \emph{this is the \textbf{first} work to successfully build an autonomous offline-to-online RL approach to enable state-of-the-art performance on device-control problems.}

\vspace{-0.2cm}
\section{Related Work}
\vspace{-0.2cm}

\textbf{Multi-modal digital agents.} In contrast to language-only agents that largely interact with both text or code inputs and outputs~\citep{toolformer, agenttuning, Chen2023FireActTL, qin2023toolllm,intercode, agentbench, jimenez2024swebench}, training multi-modal agents capable of controlling devices presents different challenges: first, device control is done directly at the pixel-level and in a coordinate-based action space, instead of natural language~\citep{aitw, xie2024osworld} that LLM is most familiar with, and second, the ecosystem of a device and the Internet tends to be quite stochastic and unpredictable, which is absent with high-level planning in language only.
To handle these challenges, prior work largely builds on strong proprietary VLMs~\citep{gpt4,geminiteam2024gemini1.5}, and designs complex rule-based wrappers~\citep{yang2023appagent, zhang2024ufo, yan2023gpt4v, zhang2024android} to enhance the visual grounding capabilities of VLMs in GUI interfaces and convert text output into pixel interactions. However, without any form of fine-tuning, this limits the room for possible performance improvement~\citep{xie2024osworld, yang2023appagent, agenttuning, Chen2023FireActTL, zhai2024fine}, especially when pre-training corpora only present limited action-labeled data. A separate line of work fine-tunes VLMs with demonstration data~\citep{lee2024benchmarking, kapoor2024omniact, hong2023cogagent, zhang2023look} via imitation learning, but maximizing single-step accuracy from stale demonstrations without accounting for consequences of these actions in subsequent steps may lead to poor solutions amidst stochasticity~\citep{pan2024autonomous}, as agents trained in such ways will struggle to recover from out-of-distribution states not included in the demonstration data~\citep{ghosh2021generalization,jiang2024importance}. 
The third category, and perhaps the closest to us, are works that run filtered imitation learning on autonomously-collected data to directly maximize the episode success rate~\citep{pan2024autonomous, lai2024autowebglm}. In contrast, \emph{ours is the first work to scale autonomous, offline-to-online RL} for device control, producing an agent that outperforms prior agents built via imitation. Even when compared to prior work running on-policy RL in simplified web navigation settings (MiniWob++~\citep{pmlr-v70-shi17a, humphreys2022datadriven}), our approach is 1000x more sample efficient (around 1e3 trajectories compared to around 1e6 trajectories), and operates in real-world GUI navigation tasks.

\textbf{Environments for device control agents.} Recent works have introduced simulated environments for building device control agents~\citep{webshop, Zhou2023WebArenaAR, koh2024visualwebarena, zhang2024mmina, drouin2024workarena,xie2024osworld}. However, these environments are primarily designed for evaluation, and present only a limited range of tasks within fully deterministic and stationary settings, infeasible for acquiring a diverse repertoire of skills needed for device control. Alternatively, other works use environments with a greater diversity of tasks~\citep{webshop, pmlr-v70-shi17a}, but these environments often oversimplify the task complexity, thus failing to transfer to in-the-wild settings. Coversely, our training environment utilizes autonomous evaluation~\citep{pan2024autonomous} with Gemini 1.5 Pro~\citep{geminiteam2024gemini1.5} to support diverse, open-ended tasks on parallel \emph{actual} Android devices, at full scale unlike prior environments. This also contrasts other prior works that use single-threaded Android emulators~\citep{pan2024autonomous, ToyamaEtAl2021AndroidEnv, lee2024benchmarking} and thus inefficient for support online RL at scale.

\textbf{Reinforcement learning for LLM/VLMs.} The majority of prior research employing RL for foundation models concentrates on tasks that must be solved in a single turn, such as preference optimization~\citep{Ouyang2022TrainingLM, ziegler2019finetuning, casper2023open} or reasoning~\citep{pang2024iterative}. However, optimizing for single-turn interaction from expert demonstrations may result in sub-optimal strategies for multi-step problems~\citep{zhou2024archer, ilql, chai}, especially amidst a high degree of stochasticity or non-stationarity. Therefore, we focus on building multi-turn RL algorithms that can learn from sub-optimal, online interaction data in this work. While prior works have developed value-based RL algorithms for LLMs~\citep{chai, ilql,lmrl, zhou2024archer, zhai2024fine}, they typically require maintaining multiple models such as Q-networks, value-networks, and policy networks, along with their delayed target counterparts, and can be subjective to slow convergence and sensitivity to choices of hyper-parameters. In contrast, we focus on identifying the key design choices for instantiating a simple yet effective RL algorithm for practitioners to incorporate to substantially improve full-scale Android device control. Our approach can serve as a base model for future research.

\vspace{-0.3cm}
\section{Problem Setup and Preliminaries}
\vspace{-0.3cm}


\textbf{Problem formulation.} We are interested in pixel-based interaction with virtual devices. We scope our study in the control of Android devices: this is already significantly more challenging and more general than previous learning-based environments that focus solely on web navigation~\citep{koh2024visualwebarena, Zhou2023WebArenaAR, drouin2024workarena}, where the web browser itself is merely one application within our broader environment, and link-based device controls~\citep{yang2023appagent, zhang2024ufo} are inadequate for tasks like games that do not support link inputs.

\begin{figure}[t]
     \centering
    \includegraphics[width=0.98\textwidth]{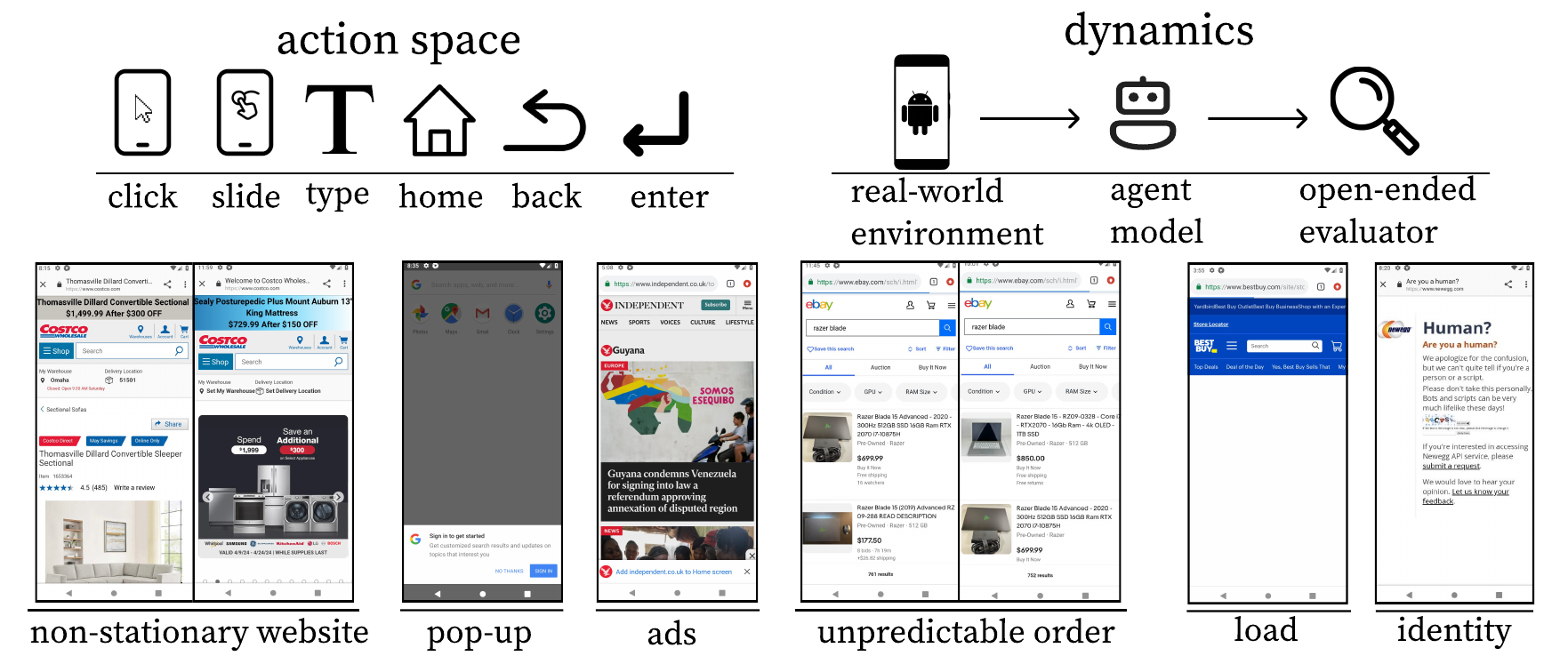}
    \caption{\footnotesize{\textbf{Environment details. }\textit{Top:} actions space and dynamics of the environment. \textit{Bottom:} examples of the read-world non-stationarity and dynamism of the environment.}}
    \label{fig:env}
    \vspace{-0.2cm}
\end{figure}

Each episode begins with the emulator initialized to the home screen. Subsequently, a task is selected from a predefined set of language instructions, some examples of which are shown in Appendix~\ref{app:env_details}. An agent is then tasked with manipulating the emulator to fulfill this instruction. At each time step, the agent receives a screenshot of the current screen as the observation. Following the action space in prior literature~\citep{aitw}, the available actions include tapping and sliding based on normalized $(x,y)$ coordinates (ranging from 0 to 1 relative to the screen dimensions), typing text strings of variable length, and pressing special buttons such as HOME, BACK, and ENTER, as illustrated in Figure~\ref{fig:env}. Our train and test instructions comes from General and Web Shopping subsets in AitW~\citep{aitw}. These tasks consist of information-gathering tasks like ``What's on the menu of In-n-Out?'', and shopping tasks on the web like ``Go to newegg.com, search for razer kraken, and select the first entry''. 

\begin{wrapfigure}{r}{0.5\textwidth}
    \centering
    \vspace{-0.5cm}
    \includegraphics[width=0.48\textwidth]{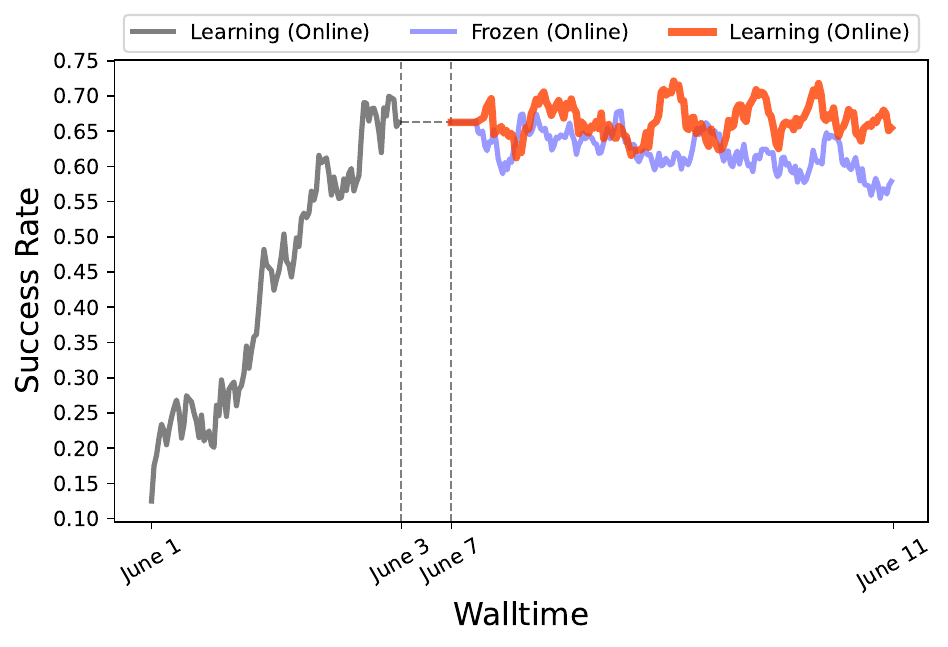}
        \caption{\footnotesize{\textbf{Performance of our approach (DigiRL) in different training modes} on the Webshop subset. When utilizing a stale checkpoint, i.e., ``frozen'' (black+\textcolor{cyan}{blue} curve) performance generally begins to degrade as time evolves, whereas autonomous online training (black+\textcolor{red}{red} curve) via DigiRL allows us to retain performance despite non-stationarity and stochasticity.}}
    \label{fig:nonstationarity}
\end{wrapfigure}

\textbf{Challenges of stochasticity.} Real-world device contrl presents unique challenges of stochasticity absent in simulated environments~\citep{Zhou2023WebArenaAR, pmlr-v70-shi17a} such as: \textbf{(1)} the non-stationarity of websites and applications, which undergo frequent updates, causing the online observations to be different from stale offline data, \textbf{(2)} various unpredictable distractors such as pop-up advertisements, login requests, and the stochastic order of search results. \textbf{(3)} technical challenges and glitches such as incomplete webpage loading or temporary access restrictions to certain sites. Examples of scenarios with such stochasticity from our experiments are shown in Figure~\ref{fig:env}. We observe that these stochastic elements pose significant challenges for pre-trained VLMs, including even those fine-tuned on device control data. As a concrete example, Figure~\ref{fig:nonstationarity} shows an experiment result that illustrates the necessity of continuously adapting the models to the non-stationarity of websites and applications. After obtaining a good checkpoint using our approach (DigiRL), that we will introduce in the next section, with autonomous data from June.1 to June.3, we compare the performance of a frozen policy and a continuously updating policy using fresh autonomous data from June.7 to June.11. We find that indeed the the performance of the frozen policy gradually degrades over time due to the changes on websites and applications, while continuous online updates plays a key role in preventing this degradation.



\textbf{Setup for reliable and scalable online RL.} As autonomous RL interleaves data collection and training, to maximize learning amidst stochasticity, it is crucial to have a real-time data collection pipeline to collect enough experience for gradient updates. While this is not possible in single-thread Android emulator environments~\citep{pan2024autonomous, ToyamaEtAl2021AndroidEnv} due to latency, we parallelize our Android emulator using appropriate error handling as discussed in Appendix~\ref{app:env_details}. In addition, the environment must provide a reward signal by judging whether the current observation indicates the agent has successfully completed the task. To generalize our \textit{evaluator} to support a wide range of tasks, we extend \citet{pan2024autonomous}'s end-to-end autonomous evaluator that does not require accessing the internal states of the emulator or human-written rules for each task. This contrasts previous works that manually write execution functions to verify the functional completeness of each task~\citep{koh2024visualwebarena, webshop, pmlr-v70-shi17a, xie2024osworld}. We adopt Gemini 1.5 Pro~\citep{geminiteam2024gemini, geminiteam2024gemini1.5} as the backbone of the autonomous evaluator. We seed this model with few-shot rollouts and the associated human-labeled success indicators to guide evaluation of novel queries. This pipeline enables a single evaluator that can evaluate all AiTW tasks. The evaluator is highly aligned with human annotations (average error rate 2.8\%), validated in Figure~\ref{fig:human_correlation}.

\vspace{-0.25cm}
\section{{\ouragentnospace}: Autonomous RL for Building a Strong Device-Control Agent}\label{sec:method}
\vspace{-0.2cm}

We now present our autonomous RL framework for training device agents. We pose the device control problem as a Markov decision process (MDP) and develop RL methods for this MDP. The core of our approach is based on a simple and scalable off-policy RL method, advantage-weighted regression (AWR)~\citep{peng2019advantageweighted}, but we make crucial modifications to handle stochasticity and highly-variable task difficulty, through the use of value functions trained with appropriate losses, and an automatic curriculum, induced by an instruction-level value function to maximize learning. 

\textbf{Device control and GUI navigation as a MDP.} 
We conceptualize device control guided by natural language instructions as a finite horizon Markov Decision Process (MDP) represented by \mbox{$\mathcal{M} = \{\mathcal{S}, \mathcal{A}, \mathcal{T}, \mu_0, \mathcal{R}, H\}$} and run policy gradient to solve this MDP. At the beginning, an initial state $s_0$ and a natural language instruction $c$ are sampled from the initial state distribution $\mu_0$. A reward of 1 is given at the end if the agent successfully fulfills the task per the evaluator, otherwise a reward of 0 is given. The trajectory terminates either when the agent accomplishes the task or when the maximum allowed number of interactions $H$ is exceeded. States are represented using the last two screenshots. To explain our approach in detail, we also include several standard definitions used in reinforcement learning (RL). The Q function for a policy $\pi$ represents the expected long-term return from taking a specific action at the current step and then following policy $\pi$ thereafter: {$Q^\pi(s_h,a_h, c) = \mathbb{E}_{\pi} \left[\sum_{t=h}^H r(s_{t}, a_{t}, c)\right]$}. The value function $V^\pi(s_h, c)$ is calculated by averaging the Q-value, $Q^\pi(s_h, a_h, c)$, over actions $a_h$ drawn from the policy $\pi$. The advantage $A^\pi(s_h,a_h, c)$ for a state-action pair is computed by subtracting the state's value under the policy from its Q-value: $A^\pi(s_h,a_h, c) = Q^\pi(s_h, a_h, c) - V^\pi(s_h, c)$.



\vspace{-0.2cm}
\subsection{Backbone of Our Approach: Off-Policy RL via Advantage-Weighted Regression}
\vspace{-0.2cm}

The starting point we choose to build our approach on is the advantage-weighted regression (AWR) algorithm~\citep{peng2019advantageweighted}, which says that we can improve the policy reliably by regressing the policy towards exponentiated advantages induced by the reward function, as a proxy for optimizing the policy gradient while staying close to the previous policy~\citep{Kakade2002ApproximatelyOA, ppo, DBLP:journals/corr/SchulmanLMJA15}:
\begin{equation}
    \mathrm{arg\,max}_{\pi}~ \mathbb{E}_{\nu} \left[\log \pi(a|s,c) \cdot \exp\left({A(s, a, c)/\beta} \right) \right],\label{eq:awr}
\end{equation}
for some positive parameter $\beta$ and the distribution of past experience $\nu$, and $A(s, a, c)$ denotes the advantage of a state-action pair $(s, a)$ given a context $c$. To avoid tuning the hyperparameter $\beta$, we consider an alternative that does ``hard filtering'' on the advantages instead of computing $\exp(A)$, similar to prior works~\citep{DBLP:journals/corr/abs-2006-09359,wang2021critic}. This leads to the following loss function for fine-tuning the model:
\begin{equation}
    \mathcal{L}(\pi) = -\mathbb{E}_{\text{filter}(\nu)}[\log \pi(a|s,c)].\label{eq:actor_loss}
\end{equation}
Typically, these advantages are computed by running Monte-Carlo (MC) rollouts in the environment to estimate the value of a given state-action pair, and subtracting from it an estimate of the value of the state given by a learned value estimator alone. However, this approach is likely to produce high-variance advantages given the stochasticity of the device eco-system that affects MC rollouts. 


\vspace{-0.2cm}
\subsection{Obtaining Reliable Advantage Estimates from Doubly-Robust Estimators}
\vspace{-0.2cm}

To reliably identify \textit{advantageous} actions given significant environment stochasticity, we construct a per-step advantage estimator, inspired by doubly-robust estimators~\citep{doubleq, schulman2018highdimensional}:
\begin{equation}
\resizebox{\textwidth}{!}{
    $A^{\text{step}}(s_h, a_h, c) := \lambda^{H-h}r(s_H, a_H, c) + (1- \lambda^{H-h}r(s_H, a_H, c))(V^{\text{step}}(s_{h+1}, c) + r(s_h, a_h, c) - V^{\text{step}}(s_h, c))$,\label{eq:doubly}}
\end{equation}
where $\lambda$ is a weighting hyper-parameter. This construction of the advantage estimator is a simplified version of Generalized Advantage Estimation (GAE)~\citep{schulman2018highdimensional} using only the next-step advantage estimator and final-step advantage estimator as there are no intermediate rewards in our problem. This construction balances an advantage estimator with higher variance Monte-Carlo estimates $\lambda^{H-h}r(s_H, a_H, c)$ (due to stochasticity) and an estimator with higher bias $V^{\text{step}}(s_{h+1}, c) + r(s_h, a_h, c) - V^{\text{step}}(s_h, c)$ (due to imperfect fitting of the value function). We observed that combining both high-variance and high-bias estimators gave us a sweet-spot in terms of performance. To implement the step-level hard filtering, we simply threshold this doubly robust estimator as $A^{\text{step}}(s_h, a_h, c) > 1/H$ to decide which actions progress towards the goal.

\vspace{-0.2cm}
\subsection{Automatic Curriculum using an Instruction-Level Value Function}
\vspace{-0.2cm}

While the AWR update (Equation \ref{eq:awr}) coupled with a robust advantage estimator (Equation~\ref{eq:doubly}) is likely sufficient on standard RL tasks, we did not find it to be effective enough for device control in preliminary experiments. Often this was the case because the task set presents tasks with highly-variable difficulties that collecting more data on tasks that the agent was already proficient at affected sample efficieny negatively. In contrast, maximal learning signal can be derived by experiencing the most informative tasks for the agent during training. To this end, we design an instruction-level value function $V^{\text{instruct}}(c)$ to evaluate if a given rollout can provide an effective learning signal:
\begin{align}
    A^{\text{instruct}}(s_h, a_h, c) := {\textstyle\sum}_{t=h}^{H}r(s_t, a_t, c) - V^{\text{instruct}}(c) = r(s_H, a_H, c) - V^{\text{instruct}}(c), \label{eq:instruct}
\end{align}
where $\sum_{t=h}^{H}r(s_t, a_t, c)$ is a Monte-Carlo estimator of $Q(s_h, a_h, c)$. The equality holds because the MDP formulation only provides rewards at the end of a rollout. Intuitively, if a rollout attains a high value of $A^{\text{instruct}}(s_h, a_h, c)$, it means the value function $V^\text{instruct}$ is small. Therefore, this rollout represents a valuable experience of the agent accomplishing a difficult task, and thus should be prioritized, akin to ideas pertaining to prioritized experience~\citep{schaul2016prioritized} or level replay~\citep{plr}. When training the actor with a buffer of historical off-policy data, we first perform a filtering step to identify the top-$p$ datapoints with highest $A^{\text{instruct}}(s_h, a_h, c)$.  Then, we use it for AWR (Equation~\ref{eq:awr}) with the doubly-robust advantage estimator (Equation~\ref{eq:doubly}).

\textbf{Implementation details.} Inspired by the findings in some recent works~\citep{farebrother2024stop, kumar2023offline} that modern deep learning architectures like transformers~\citep{vaswani2023attention} are better trained with cross-entropy losses instead of mean-squared losses, we utilize a cross-entropy objective based on the Monte-Carlo estimate of the trajectory reward for training both of our value functions:

{\vspace{-0.25cm} \small
\begin{align}
    \mathcal{L}(V^{\text{traj}}) &= -\mathbb{E}_{\nu}[r(s_H, a_H, c)\log V^{\text{traj}}(c) + (1 - r(s_H, a_H, c))\log ( 1 - V^{\text{traj}}(c))], \label{equ:train-traj-level} \\
    \mathcal{L}(V^{\text{step}}) &= -\mathbb{E}_{\nu}[r(s_H, a_H, c)\log V^{\text{step}}(s_h, a_h, c) + (1 - r(s_H, a_H, c))\log ( 1 - V^{\text{step}}(s_h, a_h, c))]. \label{equ:train-step-level}
\end{align}
}

\begin{figure}[t]
     \centering
    \includegraphics[width=0.99\textwidth]{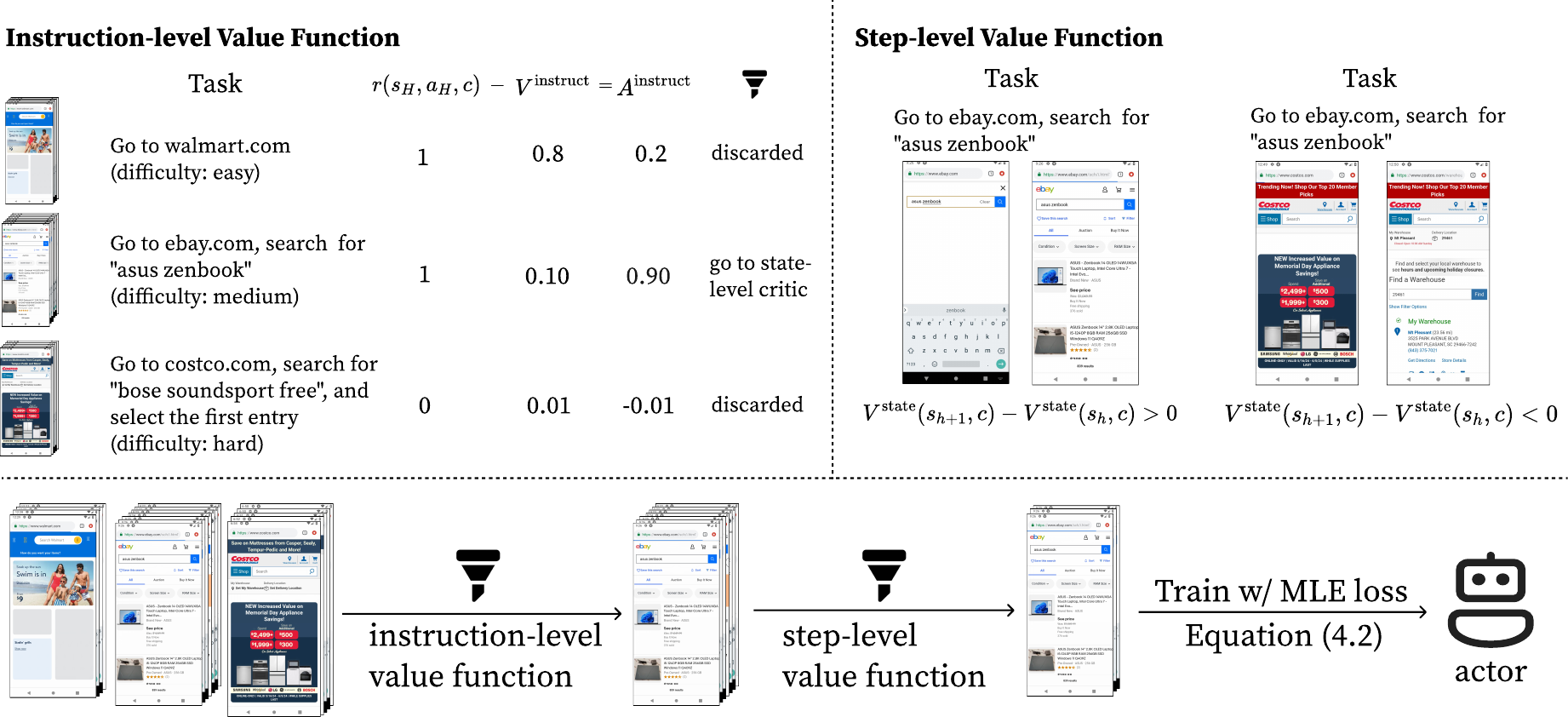}
    \caption{\footnotesize{\textbf{Algorithm visualization.} The two value function are first trained with original distribution of collected trajectories according to~\Cref{equ:train-traj-level} and~\Cref{equ:train-step-level}, then used to filter the trajectories for training the actor. We use the MLE loss (Maximum Likelihood Estimation loss) to train the actor.}}
    \label{fig:algo}
\end{figure}

\textbf{Final algorithm.} The final practical algorithm is shown in~\Cref{fig:algo}. The instruction-level value function estimates the values of the trajectories, which is trained with loss shown in~\Cref{equ:train-traj-level}. The step-level value function estimates the values of states, which is trained with loss shown in~\Cref{equ:train-step-level}. When training the actor, we first filter out trajectories and states using the value functions as shown in~\Cref{eq:instruct} and~\Cref{eq:doubly}, then train the actor with the MLE loss shown in \Cref{eq:actor_loss} on the filtered data.

\vspace{-0.2cm}
\section{Experimental Evaluation}
\vspace{-0.2cm}

The goal of our experiments is to evaluate the performance of \ouragentnospace~ on challenging Android device control problems. Specifically, we are interested in understanding if \ouragentnospace~ can produce agents that can effectively learn from autonomous interaction, while still being able to utilize offline data for learning. To this end, we perform a comparative analysis of DigiRL against several prior approaches, including state-of-the-art agents in Section~\ref{sec:main-results}. We also perform several ablation experiments to understand the necessity and sufficiency of various components of our approach in Section~\ref{sec:analysis}.

\textbf{Baselines and comparisons.} We compare \ouragent with: \textbf{(a)} state-of-the-art agents built around proprietary VLMs, with the use of several prompting and retrieval-style techniques; \textbf{(b)} running imitation learning on static human demonstrations with the same instruction distribution, and \textbf{(c)}a filtered BC approach~\citep{pan2024autonomous}.
For proprietary VLMs, we evaluate \textbf{GPT-4V}~\citep{gpt4} and \textbf{Gemini 1.5 Pro}~\citep{geminiteam2024gemini1.5} both zero-shot and when augmented with carefully-designed prompts. For the zero-shot setting, we use the prompt from \citet{yang2023appagent} and augment the observation with Set-of-Marks~\citep{zheng2024gpt4vision}. Set-of-Marks overlays a number for each interactable element over the screenshot, so that a VLM can directly output the number of the element to interact with in plain text instead of attempting to calculate pixel coordinates, which is typically significantly harder.
We also compare with AppAgent~\citep{yang2023appagent}, which first prompts the VLM to explore the environment, and appends the experience collected to the test-time prompt. We also compare with two state-of-the-art fine-tuning methods for Android device control: \textbf{AutoUI} (specifically AutoUI-Base~\citep{zhang2023look}) and \textbf{CogAgent}~\citep{hong2023cogagent}. AutoUI-Base uses an LM with 200M parameters, and a a vision encoder with 1.1B parameters. CogAgent has 11B parameters for its vision encoder and 7B for its LM. The supervised training corpus for both AutoUI-Base and CogAgent contains AitW, including the instruction set and the emulator configuration we use.

\textbf{Base VLM and offline dataset.} Both \textbf{Filtered BC} and \textbf{\ouragentnospace} use trained AutoUI-Base checkpoints with the image encoder frozen. The instruction and step-level value functions for \ouragent employ this same frozen image encoder. The visual features output from the encoder are concatenated with instruction features derived from RoBERTa~\citep{roberta}. A two-layer MLP is then used to predict the value function. In the offline phase, the offline dataset is collected by rolling out the initial AutoUI-Base supervised trained checkpoint as policy. For fair comparisons, we keep the number of offline data collected in the pure offline training roughly the same as the total number of data collected in the offline-to-online training. Due to the dynamic nature of the Internet-device eco-system, our offline data was stale by the time we were able to run our offline-to-online experiments, and this presented additional challenge in offline-to-online learning. In both General and Web Shopping subsets, offline experiments make use of around 1500 trajectories while offline-to-online experiments start with around 500 offline trajectories and update with another 1000 online trajectories. In the offline phase, \ouragent skips instruction-level filtering and instead trains the actor with all successful trajectories to make full use of the offline data. See a detailed breakdown of our dataset in Appendix~\ref{app:env_details}.


\begin{table}[!t]
    \centering
    \small
    \setlength{\tabcolsep}{5.0pt}
        \begin{tabular}{ccccccc}
            \toprule
            &&& \multicolumn{2}{c}{\textbf{AitW General}} & \multicolumn{2}{c}{\textbf{AitW Web Shopping}} \\ 
            \cmidrule(lr){4-5} \cmidrule(lr){6-7}
            &&& \texttt{Train} &\texttt{Test} & \texttt{Train} &  \texttt{Test}\\ 
            \midrule
            \multirow{4}{*}{\textbf{Prompting}} & \multirow{2}{*}{\textsc{Set-Of-Marks}} & GPT-4V & \color{gray} 5.2 & $13.5$ & \color{gray} 3.1 & $8.3$ \\
            && Gemini 1.5 Pro &  \color{gray} $32.3$ & $16.7$  & \color{gray} $6.3$ & $11.5$ \\ 
            \cdashline{2-7}
            &\multirow{2}{*}{ \begin{tabular}{@{}c@{}} \textsc{AppAgent}\end{tabular}} & GPT-4V & \color{gray}  $13.5$ & $17.7$ &\color{gray} $12.5$ & $8.3$\\
             && Gemini 1.5 Pro & \color{gray} $14.6$ & $16.7$ &      
             \color{gray} $5.2$ & $8.3$ \\
            \midrule
            \multirow{6}{*}{\textbf{Learning}} & \multirow{2}{*}{\begin{tabular}{@{}c@{}}\textsc{Supervised} \\ \textsc{Training}\end{tabular}}& CogAgent & \color{gray} $25.0$ & $25.0$ & \color{gray} $31.3$ & $38.5$\\
            && AutoUI & \color{gray} $12.5$ & $14.6$ & \color{gray} $14.6$ & $17.7$ \\
            \cdashline{2-7}
            & \multirow{2}{*}{\textsc{Offline}}
            & Filtered BC  & \color{gray} $51.7$ \scriptsize{$\pm\ 5.4$} & $50.7$ \scriptsize{$\pm\ 1.8$} & \color{gray} $44.7$ \scriptsize{$\pm\ 1.6$} & $45.8$ \scriptsize{$\pm\ 0.9$} \\
            && \textbf{Ours} & \color{gray} $46.9$ \scriptsize{$\pm\ 5.6$} & $62.8$ \scriptsize{$\pm\ 1.0$}  & \color{gray} $39.3$ \scriptsize{$\pm\ 6.0$} & $45.8$ \scriptsize{$\pm\ 6.6$} \\
            \cdashline{2-7}
            &\multirow{2}{*}{\textsc{Off-to-On}} & Filtered BC  & \color{gray} $53.5$ \scriptsize{$ \pm\ 0.8$} & $61.5$ \scriptsize{$\pm\ 1.1$} & \color{gray} $53.6$ \scriptsize{$\pm\ 4.7$} & $57.8$ \scriptsize{$\pm\ 2.6$} \\
            && \textbf{Ours} & \color{gray} $\mathbf{63.5}$ \scriptsize{$\pm\ \mathbf{0.0}$} & $\mathbf{71.9}$ \scriptsize{$\pm\ \mathbf{1.1}$} & \color{gray} $\mathbf{68.2}$ \scriptsize{$\pm\ \mathbf{6.8}$} & $\mathbf{67.2}$ \scriptsize{$\pm\ \mathbf{1.5}$} \\
            \bottomrule
        \end{tabular}
        \caption{\footnotesize{\textbf{Main comparisons of different agents across various settings.} Each offline experiment is repeated three times and the mean and standard deviation are reported. Each online experiment is repeated two times. Results are evaluated with our autonomous evaluator with the first 96 instructions in the train and test set. Correlation of our correlation
        and human judgements can be found in Figure~\ref{fig:human_correlation}.}}
        \label{tab:main-table}
        \vspace{-0.35cm}
\end{table}


\vspace{-0.2cm}
\subsection{Main Results} \label{sec:main-results}
\vspace{-0.2cm}

\begin{figure}[t]
     \centering
    \includegraphics[width=0.98\textwidth]{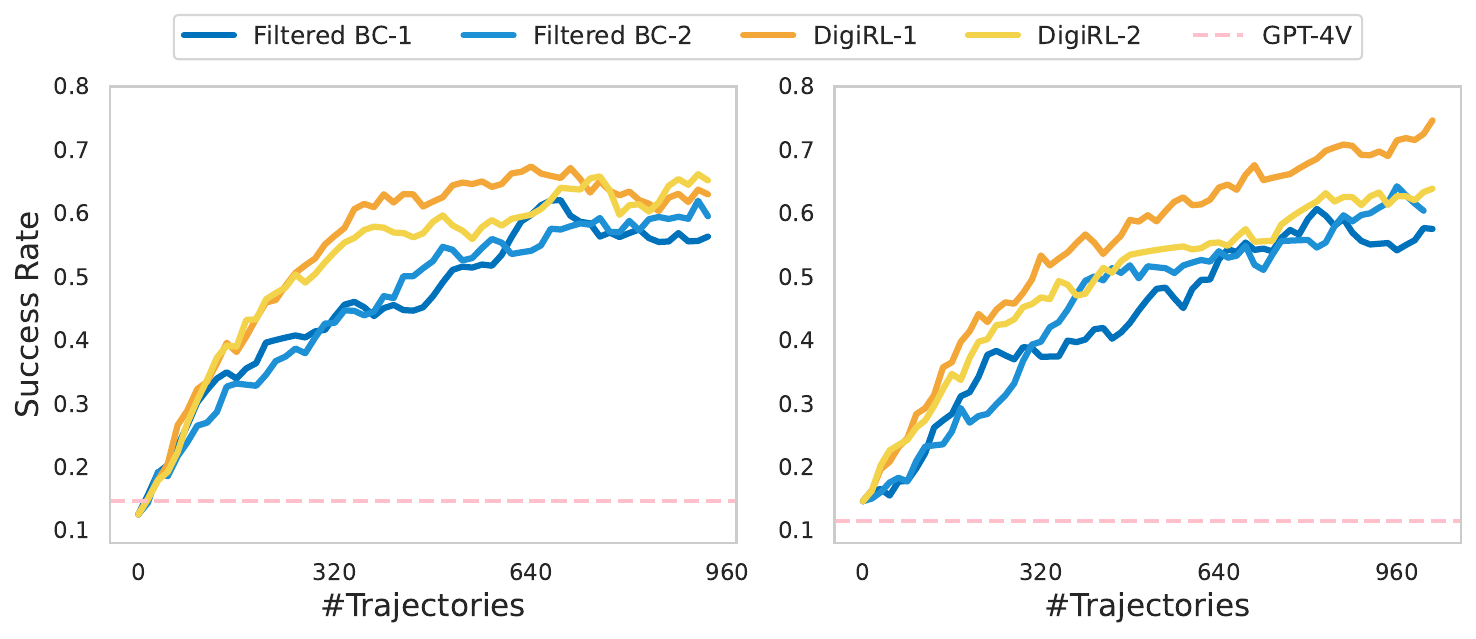}
    \caption{\footnotesize{\textbf{Offline-to-online training curves for Filtered BC and \ouragentnospace.} Curves are smoothed with exponential weighting over the x-axis. \textit{Left:} AitW General. \textit{Right:} AitW Web Shopping. Two runs for each model are started on two different dates with at least two days apart. Observe that \ouragentnospace~ is able to improve faster with a fewer number of samples. Since the data collection frequency is the bottleneck, these performance trends directly reflect performance trends against wall-clock time as well.}}
    \label{fig:online-main-results}
    \vspace{-0.3cm}
\end{figure}

Our main results are summarized in Table~\ref{tab:main-table} and Figure~\ref{fig:online-main-results}. We find that on both AitW General and AitW Web Shopping subsets, the agent trained via \ouragentnospace~ significantly outperforms prior state-of-the-art methods based on prompting and retrieval (AppAgent + GPT-4V/Gemini 1.5 Pro) or training on static demonstrations (CogAgent and AutoUI), by a large margin with more than \textbf{49.5\% absolute improvement} (from 17.7\% to 71.9\% on the General subset and from 17.7\% to 67.2\% on the Web Shopping subset). Notably, this improvement from \ouragent is realized \emph{fully autonomously without making use of human supervision} (e.g. manually labeled rollouts or hand-written verifiers).

\begin{figure}[t]
    \centering
    \includegraphics[width=0.9\textwidth]{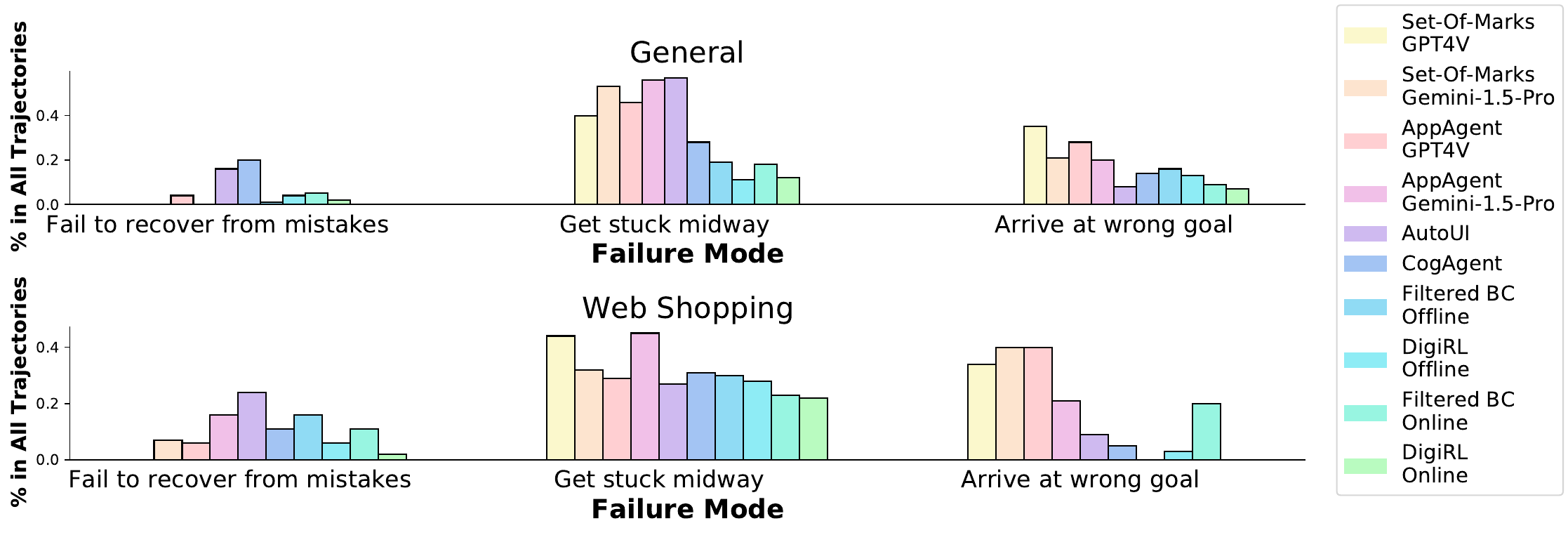}
  \caption{\footnotesize{\textbf{Failure modes for each approach} on both the AiTW General and Web Shopping subsets. We found that the failure mode RL training is most effective at reducing compared to model supervised trained on human data is ``Fail to recover from mistakes''. A more fine-grained decomposition can be found in Appendix~\ref{app:fine-grained}. }}
  \label{fig:error_analysis}
\end{figure}

\textbf{Are inference-time prompting and retrieval techniques or supervised training enough for device control?} Delving into Table~\ref{tab:main-table}, we  observe that off-the-shelf proprietary VLMs, even when supplemented with the set-of-marks mechanism, do not attain satisfactory performance: both GPT-4V and Gemini 1.5 Pro achieve success rates under 20\%. One possible cause could be the under-representation of Android device data in the pre-training data. Moreover, inference-time adaptation strategies such as AppAgent~\citep{yang2023appagent} show minimal improvement, with gains not exceeding 5\% for either model. All this evidence suggests a limited scope for improvement without fine-tuning of some sort. As illustrated in Figure~\ref{fig:error_analysis}, the primary failures of these VLMs stem from hallucinatory reasoning that lead the VLMs to land on a relevant but wrong page. This suggests that while state-of-the-art VLMs excel at reasoning problems in  code and math, their reliability in less-familiar domains, such as device control, remains inadequate. For example, for the instruction ``Go to newegg.com, search for alienware area 51, and select the first entry'', a GPT-4V based agent erroneously searched ``alien area 51 ebay'' in Google.com and decided that it had made progress towards the task (Figure~\ref{fig:gpt4_search_failure}).

Training on domain-specific human demonstrations, however, does boost performance, allowing the smaller, specialized VLM, AutoUI with 1.5 billion parameters, to match or surpass the larger, generalist VLMs like GPT-4V and Gemini 1.5 Pro. Nonetheless, this supervised imitation learning approach still fall short, with success rates on both subsets remaining below 20\%. This shortcoming is not fundamentally addressed via enhancements in model scale or architecture, as evidenced by CogAgent~\citep{hong2023cogagent}, with 18 billion parameters still achieving performances below 40\% success rate. As depicted in Figure~\ref{fig:error_analysis}, a predominant failure mode for these agents is an inability to rectify their own errors. An example trajectory that we observed is that for the instruction ``what's on the menu of In-n-Out'', the agent accidentally activated the voice input button, and failed to quit that page until the step limit. In contrast, DigiRL is able to recover from the errors more efficiently(~\Cref{app:error-recovery}).


\textbf{Comparison of different RL approaches.} In Table~\ref{tab:main-table} and Figure~\ref{fig:online-main-results}, we present a comparative analysis of various autonomous approaches. Notably, both offline and offline-to-online configurations demonstrate that our RL approach, when augmented with a continuous stream of autonomous interaction data and reward feedback, substantially improves performance. This improvement is evident from an increase in the success rate from under 20\% to over 40\%, as the agent learns to adapt to stochastic and non-stationary device interfaces. Moreover, although the total sample sizes for offline and offline-to-online settings are equivalent, the top-performing offline-to-online algorithm markedly surpasses its offline counterpart (75\% versus 62.8\% on the General subset). This highlights the efficacy of autonomous environment interaction, and establishes the efficacy of DigiRL in learning from such uncurated, sub-optimal data. Lastly, \ouragent consistently outperforms the state-of-the-art alternative, Filtered BC, across both the General and Web Shopping subsets, improving from 61.5\% to 71.9\% and 57.8\% to 61.4\%, respectively, highlighting DigiRL's performance and efficiency. 




\vspace{-0.2cm}
\subsection{Analysis and Ablations}\label{sec:analysis}
\vspace{-0.2cm}

\textbf{Failure modes analysis.} 
We conduct an additional user study to annotate the failure modes for each agent as shown in Figure~\ref{fig:error_analysis}, and a more fine-grained breakdown can be found in Appendix~\ref{app:fine-grained}. At a high level, we classify the major failure modes of all agents into the following three categories: \textbf{(1) \textit{Failure to recover from mistakes}} refers to the scenario where the agent made a mistake that led it to states from which it failed to quickly recover and resume the task, such as a wrong search page. \textbf{(2) \textit{Getting stuck midway}} refers to the failure mode where the agent gets distracted on the right track to completing the instruction and as a result fails to accomplish the task. For example, failing to click on the right link or failing to search after typing the key words. \textbf{(3) \textit{Arriving at wrong goal}} refers to the failure mode where the agent arrives at a wrong page and mistakenly thinks that it had completed the task. For e.g, the agent finds a macbook on costco.com instead of finding a macbook on ebay.com.

While all the types of failure modes benefit from offline and offline-to-online RL training as shown in~\Cref{fig:error_analysis}, the most consistent and significant reduction is probably for the failure mode of failing to recover from mistakes. This is because while pre-trained models, generating plausible future tokens, can get distracted by the dynamic nature of the environment and, as a result, encounter at never-before-seen states. With no clue of how to escape such states, these methods are unable to recover and fail to solve the task. In contrast, by training on autonomously-collected rollouts, our agent DigiRL is able to learn from its own mistakes and reduces failures to recover over training.

\begin{figure}[!t]
     \centering
    \includegraphics[width=1.0\textwidth]{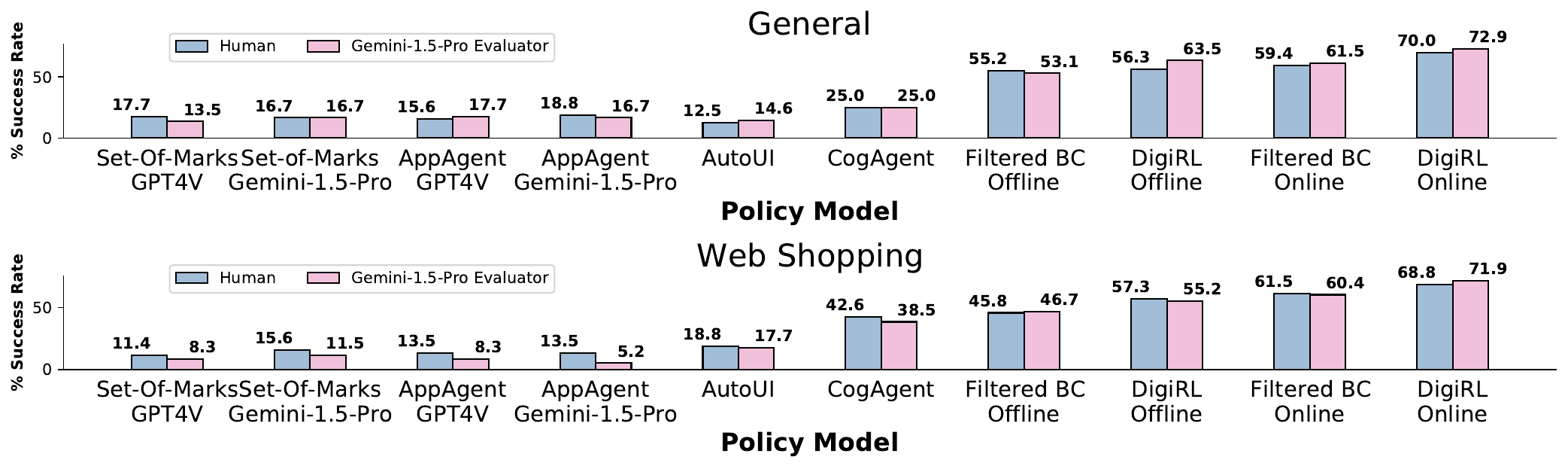}
    \vspace{-0.1cm}
    \caption{\footnotesize{\textbf{Correlation between our autonomous evaluator and human judgements for all policy models} on General and Web Shopping subsets. For repeated offline and online runs, we report the correlation results for the run with the highest autonomous evaluation success rate.}}
    \label{fig:human_correlation}
    \vspace{-0.5cm}
\end{figure}

\begin{figure}[!t]
     \centering
     \begin{subfigure}[b]{0.55\textwidth}
         \centering
    \includegraphics[width=\textwidth]{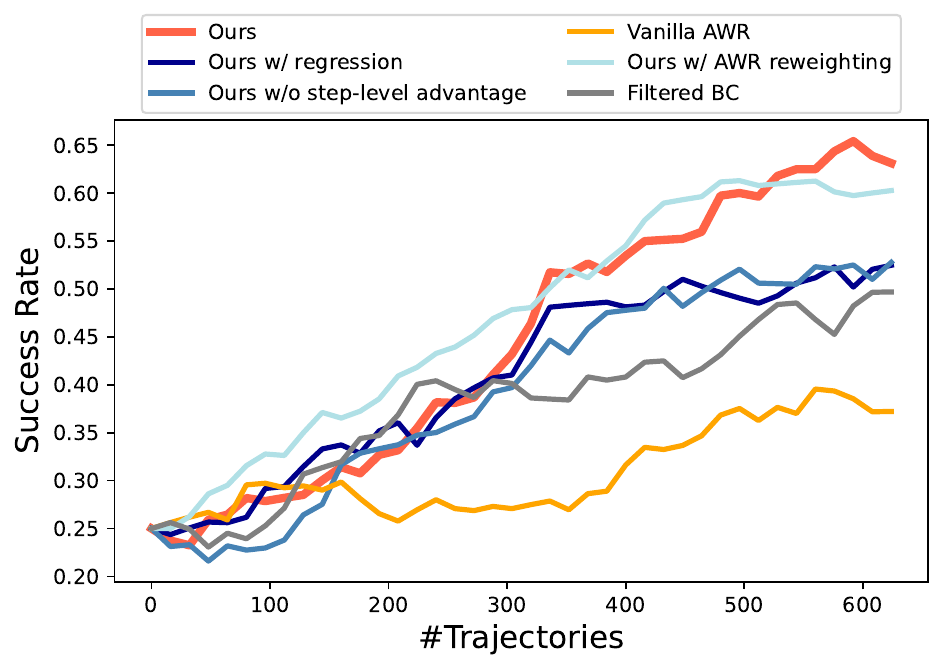}
     \end{subfigure}
    \begin{subfigure}[b]{0.39\textwidth}
         \centering
    \includegraphics[width=\textwidth]{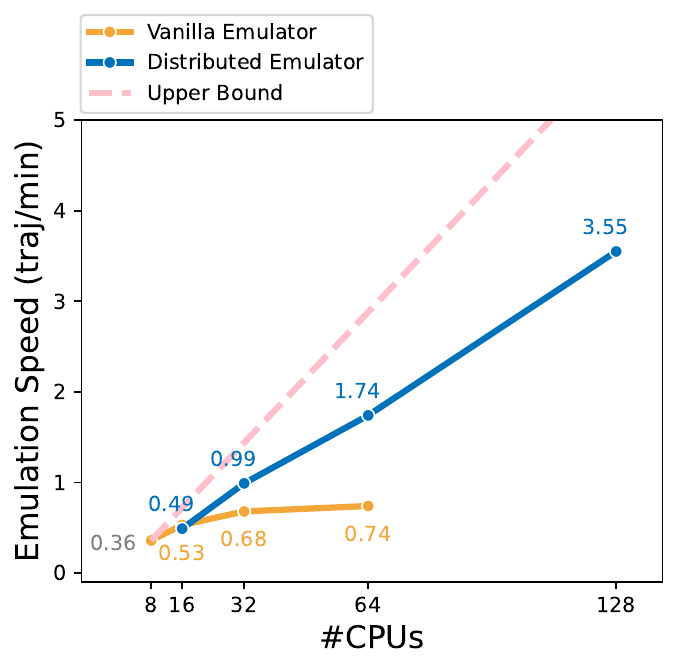}
     \end{subfigure}
        \caption{\footnotesize{\textit{Left:} \textbf{Ablation study results on the AitW Web Shopping subset.}
        \textit{Right:} \textbf{Emulation speed w.r.t number of CPUs used.} The upper bound can only achieved when there is no communication and error handling cost. Our design of distributed emulator can significantly improve the efficiency of emulation compaared to the vanilla method of running all emulations over the same instance.}}
        \label{fig:parallel-and-ablation}
        \vspace{-0.2cm}
\end{figure}

\textbf{Ablation study of each component in \ouragentnospace.} We conduct an ablation study on different components of \ouragent in Figure~\ref{fig:parallel-and-ablation} (left). We find that all the components used by our approach are necessary: \textbf{(1)} using cross-entropy for training the value functions boosts performance by around 12\% (compare Ours and Ours w/ Regression); \textbf{(2)} using step-level advantages improves efficiency by 12\% (comparing Ours and Ours w/o step-level advantage); \textbf{(3)} the use of automatic curriculum improves the speed of learning by around 25\% (comparing Ours w/o step-level advantage and Filtered BC); \textbf{(4)} Ours outperforms vanilla AWR that does not employ a doubly-robust advantage estimator or curriculum.

Additionally, we also observe no degradation in performance as a result of ``hard-filtering'', as show by nearly comparable performance of our approach and the best run of exponential filtering obtained via an extensive tuning of the temperature hyperparameter $\tau$ in na\"ive AWR (comparing Ours and Ours w/ vanilla AWR reweighting), despite simplicity of implementation in the hard filtering approach. Putting together, these choices result in a new state-of-the-art RL approach for device control.


\textbf{Evaluation of our autonomous evaluator.} 
In Figure~\ref{fig:human_correlation}, we present the findings from a user study aimed at assessing the accuracy of our autonomous evaluator. Our results indicate that the success rates reported by our automatic evaluator are remarkably consistent with those assessed by human evaluators across almost all models, with differences less than 3\%. Furthermore, we observed that evaluations on the Web Shopping subset are more precise compared to those on the General subset. This increased accuracy likely stems from the fact that tasks in the General subset are formulated in free-form language, which can introduce ambiguity, whereas the Web Shopping subset features a narrower range of language expressions, reducing potential variability.

\textbf{Speedup of emulation parallel.} The performance boost with respect to the number of worker machines is nearly linear, as demonstrated in~\Cref{fig:parallel-and-ablation} (right), where we conduct experiments that examine the scaling performance of our parallel emulator. Our distributed emulator that runs emulations across multiple servers can reliably collect data with up to 64 parallel emulators on 128 CPUs with near-linear speedup. In contrast, a naive baseline that runs all parallel emulations on the same server achieves much inferior performance (0.74 compared to 1.74 trajs/min using 64 CPUs).

\vspace{-0.2cm}
\section{Discussion and Limitations} \label{sec:limitations}
\vspace{-0.2cm}

In this paper, we propose a novel autonomous RL approach, \ouragentnospace, for training in-the-wild, multi-modal, device-control agents that establish a new state-of-the-art performance on a number of Android control tasks from Android-in-the-Wild dataset~\citep{aitw}. To achieve this, we first build a scalable and parallelizable Android environment with a robust VLM-based general-purpose evaluator that supports fast online data collection. We then develop a system for offline RL pre-training, followed by autonomous RL fine-tuning to learn via interaction, admist the stochasticity of the real-world Internet and device eco-system. Our agent achieves a 280\% improvement over the previous state-of-the-art agents (from 17.7\% to 68.2\% in terms of task success rate), including AppAgent based on GPT-4V and Gemini 1.5 Pro, and supervised trained models such as AutoUI and CogAgent.

Due to computational limitations, and despite the fact that the parallel emulator and autonomous evaluator can be easily extended to complicated tasks, our agent is trained only with tasks from AitW instead of a all possible tasks on the device. Our design of the DigiRL algorithm aims for maximal implementation simplicity, so we hope that our approach to serve as a base algorithm for future research to build on, including algorithmic research as well as expanding the space of tasks.

\vspace{-0.2cm}
\section*{Acknowledgements}
\vspace{-0.2cm}

We thank Yi Su, Izzedin Gur, Xinyang Geng, and Sandra Faust for feedback on an earlier version of this paper and for informative discussions. This work is supported by NSF IIS-2246811 and ONR N00014-21-1-2838, and Gemini 1.5 Pro credit donations for academic use and cloud resources from Google Cloud. 


\bibliography{neurips_2023}

\begin{thebibliography}{58}
\providecommand{\natexlab}[1]{#1}
\providecommand{\url}[1]{\texttt{#1}}
\expandafter\ifx\csname urlstyle\endcsname\relax
  \providecommand{\doi}[1]{doi: #1}\else
  \providecommand{\doi}{doi: \begingroup \urlstyle{rm}\Url}\fi

\bibitem[Abdulhai et~al.(2023)Abdulhai, White, Snell, Sun, Hong, Zhai, Xu, and
  Levine]{lmrl}
Marwa Abdulhai, Isadora White, Charlie Snell, Charles Sun, Joey Hong, Yuexiang
  Zhai, Kelvin Xu, and Sergey Levine.
\newblock Lmrl gym: Benchmarks for multi-turn reinforcement learning with
  language models, 2023.

\bibitem[Casper et~al.(2023)Casper, Davies, Shi, Gilbert, Scheurer, Rando,
  Freedman, Korbak, Lindner, Freire, Wang, Marks, Segerie, Carroll, Peng,
  Christoffersen, Damani, Slocum, Anwar, Siththaranjan, Nadeau, Michaud, Pfau,
  Krasheninnikov, Chen, Langosco, Hase, Bıyık, Dragan, Krueger, Sadigh, and
  Hadfield-Menell]{casper2023open}
Stephen Casper, Xander Davies, Claudia Shi, Thomas~Krendl Gilbert, Jérémy
  Scheurer, Javier Rando, Rachel Freedman, Tomasz Korbak, David Lindner, Pedro
  Freire, Tony Wang, Samuel Marks, Charbel-Raphaël Segerie, Micah Carroll,
  Andi Peng, Phillip Christoffersen, Mehul Damani, Stewart Slocum, Usman Anwar,
  Anand Siththaranjan, Max Nadeau, Eric~J. Michaud, Jacob Pfau, Dmitrii
  Krasheninnikov, Xin Chen, Lauro Langosco, Peter Hase, Erdem Bıyık, Anca
  Dragan, David Krueger, Dorsa Sadigh, and Dylan Hadfield-Menell.
\newblock Open problems and fundamental limitations of reinforcement learning
  from human feedback, 2023.

\bibitem[Chen et~al.(2023)Chen, Shu, Shareghi, Collier, Narasimhan, and
  Yao]{Chen2023FireActTL}
Baian Chen, Chang Shu, Ehsan Shareghi, Nigel Collier, Karthik Narasimhan, and
  Shunyu Yao.
\newblock Fireact: Toward language agent fine-tuning.
\newblock \emph{ArXiv}, abs/2310.05915, 2023.
\newblock URL \url{https://api.semanticscholar.org/CorpusID:263829338}.

\bibitem[Drouin et~al.(2024)Drouin, Gasse, Caccia, Laradji, Verme, Marty,
  Boisvert, Thakkar, Cappart, Vazquez, Chapados, and
  Lacoste]{drouin2024workarena}
Alexandre Drouin, Maxime Gasse, Massimo Caccia, Issam~H. Laradji, Manuel~Del
  Verme, Tom Marty, Léo Boisvert, Megh Thakkar, Quentin Cappart, David
  Vazquez, Nicolas Chapados, and Alexandre Lacoste.
\newblock Workarena: How capable are web agents at solving common knowledge
  work tasks?, 2024.

\bibitem[Farebrother et~al.(2024)Farebrother, Orbay, Vuong, Taïga, Chebotar,
  Xiao, Irpan, Levine, Castro, Faust, Kumar, and Agarwal]{farebrother2024stop}
Jesse Farebrother, Jordi Orbay, Quan Vuong, Adrien~Ali Taïga, Yevgen Chebotar,
  Ted Xiao, Alex Irpan, Sergey Levine, Pablo~Samuel Castro, Aleksandra Faust,
  Aviral Kumar, and Rishabh Agarwal.
\newblock Stop regressing: Training value functions via classification for
  scalable deep rl, 2024.

\bibitem[Gemini~Team(2024{\natexlab{a}})]{geminiteam2024gemini}
2023 Gemini~Team.
\newblock Gemini: A family of highly capable multimodal models,
  2024{\natexlab{a}}.

\bibitem[Gemini~Team(2024{\natexlab{b}})]{geminiteam2024gemini1.5}
2024 Gemini~Team.
\newblock Gemini 1.5: Unlocking multimodal understanding across millions of
  tokens of context, 2024{\natexlab{b}}.

\bibitem[Ghosh et~al.(2021)Ghosh, Rahme, Kumar, Zhang, Adams, and
  Levine]{ghosh2021generalization}
Dibya Ghosh, Jad Rahme, Aviral Kumar, Amy Zhang, Ryan~P Adams, and Sergey
  Levine.
\newblock {Why Generalization in RL is Difficult: Epistemic POMDPs and Implicit
  Partial Observability}.
\newblock \emph{NeurIPS}, 2021.

\bibitem[Hong et~al.(2023)Hong, Wang, Lv, Xu, Yu, Ji, Wang, Wang, Zhang, Li,
  Xu, Dong, Ding, and Tang]{hong2023cogagent}
Wenyi Hong, Weihan Wang, Qingsong Lv, Jiazheng Xu, Wenmeng Yu, Junhui Ji, Yan
  Wang, Zihan Wang, Yuxuan Zhang, Juanzi Li, Bin Xu, Yuxiao Dong, Ming Ding,
  and Jie Tang.
\newblock Cogagent: A visual language model for gui agents, 2023.

\bibitem[Humphreys et~al.(2022)Humphreys, Raposo, Pohlen, Thornton, Chhaparia,
  Muldal, Abramson, Georgiev, Goldin, Santoro, and
  Lillicrap]{humphreys2022datadriven}
Peter~C Humphreys, David Raposo, Toby Pohlen, Gregory Thornton, Rachita
  Chhaparia, Alistair Muldal, Josh Abramson, Petko Georgiev, Alex Goldin, Adam
  Santoro, and Timothy Lillicrap.
\newblock A data-driven approach for learning to control computers, 2022.

\bibitem[Jiang et~al.(2020)Jiang, Grefenstette, and Rockt{\"{a}}schel]{plr}
Minqi Jiang, Edward Grefenstette, and Tim Rockt{\"{a}}schel.
\newblock Prioritized level replay.
\newblock \emph{CoRR}, abs/2010.03934, 2020.
\newblock URL \url{https://arxiv.org/abs/2010.03934}.

\bibitem[Jiang et~al.(2024)Jiang, Kolter, and Raileanu]{jiang2024importance}
Yiding Jiang, J~Zico Kolter, and Roberta Raileanu.
\newblock On the importance of exploration for generalization in reinforcement
  learning.
\newblock \emph{Advances in Neural Information Processing Systems}, 36, 2024.

\bibitem[Jimenez et~al.(2024)Jimenez, Yang, Wettig, Yao, Pei, Press, and
  Narasimhan]{jimenez2024swebench}
Carlos~E. Jimenez, John Yang, Alexander Wettig, Shunyu Yao, Kexin Pei, Ofir
  Press, and Karthik Narasimhan.
\newblock Swe-bench: Can language models resolve real-world github issues?,
  2024.

\bibitem[Kakade and Langford(2002)]{Kakade2002ApproximatelyOA}
Sham~M. Kakade and John Langford.
\newblock Approximately optimal approximate reinforcement learning.
\newblock In \emph{International Conference on Machine Learning}, 2002.
\newblock URL \url{https://api.semanticscholar.org/CorpusID:31442909}.

\bibitem[Kapoor et~al.(2024)Kapoor, Butala, Russak, Koh, Kamble, Alshikh, and
  Salakhutdinov]{kapoor2024omniact}
Raghav Kapoor, Yash~Parag Butala, Melisa Russak, Jing~Yu Koh, Kiran Kamble,
  Waseem Alshikh, and Ruslan Salakhutdinov.
\newblock Omniact: A dataset and benchmark for enabling multimodal generalist
  autonomous agents for desktop and web, 2024.

\bibitem[Koh et~al.(2024)Koh, Lo, Jang, Duvvur, Lim, Huang, Neubig, Zhou,
  Salakhutdinov, and Fried]{koh2024visualwebarena}
Jing~Yu Koh, Robert Lo, Lawrence Jang, Vikram Duvvur, Ming~Chong Lim, Po-Yu
  Huang, Graham Neubig, Shuyan Zhou, Ruslan Salakhutdinov, and Daniel Fried.
\newblock Visualwebarena: Evaluating multimodal agents on realistic visual web
  tasks.
\newblock \emph{arXiv preprint arXiv:2401.13649}, 2024.

\bibitem[Kumar et~al.(2023)Kumar, Agarwal, Geng, Tucker, and
  Levine]{kumar2023offline}
Aviral Kumar, Rishabh Agarwal, Xinyang Geng, George Tucker, and Sergey Levine.
\newblock Offline q-learning on diverse multi-task data both scales and
  generalizes, 2023.

\bibitem[Lai et~al.(2024)Lai, Liu, Iong, Yao, Chen, Shen, Yu, Zhang, Zhang,
  Dong, and Tang]{lai2024autowebglm}
Hanyu Lai, Xiao Liu, Iat~Long Iong, Shuntian Yao, Yuxuan Chen, Pengbo Shen, Hao
  Yu, Hanchen Zhang, Xiaohan Zhang, Yuxiao Dong, and Jie Tang.
\newblock Autowebglm: Bootstrap and reinforce a large language model-based web
  navigating agent, 2024.

\bibitem[Lee et~al.(2024)Lee, Min, An, Kim, and Lee]{lee2024benchmarking}
Juyong Lee, Taywon Min, Minyong An, Changyeon Kim, and Kimin Lee.
\newblock Benchmarking mobile device control agents across diverse
  configurations, 2024.

\bibitem[Liu et~al.(2023)Liu, Yu, Zhang, Xu, Lei, Lai, Gu, Ding, Men, Yang,
  Zhang, Deng, Zeng, Du, Zhang, Shen, Zhang, Su, Sun, Huang, Dong, and
  Tang]{agentbench}
Xiao Liu, Hao Yu, Hanchen Zhang, Yifan Xu, Xuanyu Lei, Hanyu Lai, Yu~Gu,
  Hangliang Ding, Kaiwen Men, Kejuan Yang, Shudan Zhang, Xiang Deng, Aohan
  Zeng, Zhengxiao Du, Chenhui Zhang, Sheng Shen, Tianjun Zhang, Yu~Su, Huan
  Sun, Minlie Huang, Yuxiao Dong, and Jie Tang.
\newblock Agentbench: Evaluating llms as agents, 2023.

\bibitem[Liu et~al.(2019)Liu, Ott, Goyal, Du, Joshi, Chen, Levy, Lewis,
  Zettlemoyer, and Stoyanov]{roberta}
Yinhan Liu, Myle Ott, Naman Goyal, Jingfei Du, Mandar Joshi, Danqi Chen, Omer
  Levy, Mike Lewis, Luke Zettlemoyer, and Veselin Stoyanov.
\newblock Roberta: {A} robustly optimized {BERT} pretraining approach.
\newblock \emph{CoRR}, abs/1907.11692, 2019.
\newblock URL \url{http://arxiv.org/abs/1907.11692}.

\bibitem[Nair et~al.(2020)Nair, Dalal, Gupta, and
  Levine]{DBLP:journals/corr/abs-2006-09359}
Ashvin Nair, Murtaza Dalal, Abhishek Gupta, and Sergey Levine.
\newblock Accelerating online reinforcement learning with offline datasets.
\newblock \emph{CoRR}, abs/2006.09359, 2020.
\newblock URL \url{https://arxiv.org/abs/2006.09359}.

\bibitem[OpenAI et~al.(2019)OpenAI, Akkaya, Andrychowicz, Chociej, Litwin,
  McGrew, Petron, Paino, Plappert, Powell, Ribas, Schneider, Tezak, Tworek,
  Welinder, Weng, Yuan, Zaremba, and Zhang]{openai2019solving}
OpenAI, Ilge Akkaya, Marcin Andrychowicz, Maciek Chociej, Mateusz Litwin, Bob
  McGrew, Arthur Petron, Alex Paino, Matthias Plappert, Glenn Powell, Raphael
  Ribas, Jonas Schneider, Nikolas Tezak, Jerry Tworek, Peter Welinder, Lilian
  Weng, Qiming Yuan, Wojciech Zaremba, and Lei Zhang.
\newblock Solving rubik's cube with a robot hand, 2019.

\bibitem[OpenAI~Team(2023)]{gpt4}
2023 OpenAI~Team.
\newblock Gpt-4 technical report, 2023.

\bibitem[Ouyang et~al.(2022)Ouyang, Wu, Jiang, Almeida, Wainwright, Mishkin,
  Zhang, Agarwal, Slama, Ray, Schulman, Hilton, Kelton, Miller, Simens, Askell,
  Welinder, Christiano, Leike, and Lowe]{Ouyang2022TrainingLM}
Long Ouyang, Jeff Wu, Xu~Jiang, Diogo Almeida, Carroll~L. Wainwright, Pamela
  Mishkin, Chong Zhang, Sandhini Agarwal, Katarina Slama, Alex Ray, John
  Schulman, Jacob Hilton, Fraser Kelton, Luke~E. Miller, Maddie Simens, Amanda
  Askell, Peter Welinder, Paul~Francis Christiano, Jan Leike, and Ryan~J. Lowe.
\newblock Training language models to follow instructions with human feedback.
\newblock \emph{ArXiv}, abs/2203.02155, 2022.
\newblock URL \url{https://api.semanticscholar.org/CorpusID:246426909}.

\bibitem[Pan et~al.(2024)Pan, Zhang, Tomlin, Zhou, Levine, and
  Suhr]{pan2024autonomous}
Jiayi Pan, Yichi Zhang, Nicholas Tomlin, Yifei Zhou, Sergey Levine, and Alane
  Suhr.
\newblock Autonomous evaluation and refinement of digital agents.
\newblock \emph{arXiv preprint arXiv:2404.06474}, 2024.

\bibitem[Pang et~al.(2024)Pang, Yuan, Cho, He, Sukhbaatar, and
  Weston]{pang2024iterative}
Richard~Yuanzhe Pang, Weizhe Yuan, Kyunghyun Cho, He~He, Sainbayar Sukhbaatar,
  and Jason Weston.
\newblock Iterative reasoning preference optimization, 2024.

\bibitem[Peng et~al.(2019{\natexlab{a}})Peng, Kumar, Zhang, and Levine]{awr}
Xue~Bin Peng, Aviral Kumar, Grace Zhang, and Sergey Levine.
\newblock Advantage-weighted regression: Simple and scalable off-policy
  reinforcement learning.
\newblock \emph{CoRR}, abs/1910.00177, 2019{\natexlab{a}}.
\newblock URL \url{http://arxiv.org/abs/1910.00177}.

\bibitem[Peng et~al.(2019{\natexlab{b}})Peng, Kumar, Zhang, and
  Levine]{peng2019advantageweighted}
Xue~Bin Peng, Aviral Kumar, Grace Zhang, and Sergey Levine.
\newblock Advantage-weighted regression: Simple and scalable off-policy
  reinforcement learning, 2019{\natexlab{b}}.

\bibitem[Qin et~al.(2023)Qin, Liang, Ye, Zhu, Yan, Lu, Lin, Cong, Tang, Qian,
  Zhao, Hong, Tian, Xie, Zhou, Gerstein, Li, Liu, and Sun]{qin2023toolllm}
Yujia Qin, Shihao Liang, Yining Ye, Kunlun Zhu, Lan Yan, Yaxi Lu, Yankai Lin,
  Xin Cong, Xiangru Tang, Bill Qian, Sihan Zhao, Lauren Hong, Runchu Tian,
  Ruobing Xie, Jie Zhou, Mark Gerstein, Dahai Li, Zhiyuan Liu, and Maosong Sun.
\newblock Toolllm: Facilitating large language models to master 16000+
  real-world apis, 2023.

\bibitem[Rawles et~al.(2023)Rawles, Li, Rodriguez, Riva, and Lillicrap]{aitw}
Christopher Rawles, Alice Li, Daniel Rodriguez, Oriana Riva, and Timothy
  Lillicrap.
\newblock Android in the wild: A large-scale dataset for android device
  control.
\newblock \emph{arXiv preprint arXiv:2307.10088}, 2023.

\bibitem[Schaul et~al.(2016)Schaul, Quan, Antonoglou, and
  Silver]{schaul2016prioritized}
Tom Schaul, John Quan, Ioannis Antonoglou, and David Silver.
\newblock Prioritized experience replay, 2016.

\bibitem[Schick et~al.(2023)Schick, Dwivedi-Yu, Dessì, Raileanu, Lomeli,
  Zettlemoyer, Cancedda, and Scialom]{toolformer}
Timo Schick, Jane Dwivedi-Yu, Roberto Dessì, Roberta Raileanu, Maria Lomeli,
  Luke Zettlemoyer, Nicola Cancedda, and Thomas Scialom.
\newblock Toolformer: Language models can teach themselves to use tools, 2023.

\bibitem[Schulman et~al.(2015)Schulman, Levine, Moritz, Jordan, and
  Abbeel]{DBLP:journals/corr/SchulmanLMJA15}
John Schulman, Sergey Levine, Philipp Moritz, Michael~I. Jordan, and Pieter
  Abbeel.
\newblock Trust region policy optimization.
\newblock \emph{CoRR}, abs/1502.05477, 2015.
\newblock URL \url{http://arxiv.org/abs/1502.05477}.

\bibitem[Schulman et~al.(2017)Schulman, Wolski, Dhariwal, Radford, and
  Klimov]{ppo}
John Schulman, Filip Wolski, Prafulla Dhariwal, Alec Radford, and Oleg Klimov.
\newblock Proximal policy optimization algorithms.
\newblock \emph{CoRR}, abs/1707.06347, 2017.
\newblock URL \url{http://arxiv.org/abs/1707.06347}.

\bibitem[Schulman et~al.(2018)Schulman, Moritz, Levine, Jordan, and
  Abbeel]{schulman2018highdimensional}
John Schulman, Philipp Moritz, Sergey Levine, Michael Jordan, and Pieter
  Abbeel.
\newblock High-dimensional continuous control using generalized advantage
  estimation, 2018.

\bibitem[Shi et~al.(2017)Shi, Karpathy, Fan, Hernandez, and
  Liang]{pmlr-v70-shi17a}
Tianlin Shi, Andrej Karpathy, Linxi Fan, Jonathan Hernandez, and Percy Liang.
\newblock World of bits: An open-domain platform for web-based agents.
\newblock In Doina Precup and Yee~Whye Teh, editors, \emph{Proceedings of the
  34th International Conference on Machine Learning}, volume~70 of
  \emph{Proceedings of Machine Learning Research}, pages 3135--3144. PMLR,
  06--11 Aug 2017.
\newblock URL \url{https://proceedings.mlr.press/v70/shi17a.html}.

\bibitem[Snell et~al.(2023)Snell, Kostrikov, Su, Yang, and Levine]{ilql}
Charlie Snell, Ilya Kostrikov, Yi~Su, Mengjiao Yang, and Sergey Levine.
\newblock Offline rl for natural language generation with implicit language q
  learning, 2023.

\bibitem[Toyama et~al.(2021)Toyama, Hamel, Gergely, Comanici, Glaese, Ahmed,
  Jackson, Mourad, and Precup]{ToyamaEtAl2021AndroidEnv}
Daniel Toyama, Philippe Hamel, Anita Gergely, Gheorghe Comanici, Amelia Glaese,
  Zafarali Ahmed, Tyler Jackson, Shibl Mourad, and Doina Precup.
\newblock Androidenv: A reinforcement learning platform for android.
\newblock \emph{arXiv preprint arXiv:2105.13231}, 2021.

\bibitem[van Hasselt et~al.(2015)van Hasselt, Guez, and Silver]{doubleq}
Hado van Hasselt, Arthur Guez, and David Silver.
\newblock Deep reinforcement learning with double q-learning.
\newblock \emph{CoRR}, abs/1509.06461, 2015.
\newblock URL \url{http://arxiv.org/abs/1509.06461}.

\bibitem[Vaswani et~al.(2023)Vaswani, Shazeer, Parmar, Uszkoreit, Jones, Gomez,
  Kaiser, and Polosukhin]{vaswani2023attention}
Ashish Vaswani, Noam Shazeer, Niki Parmar, Jakob Uszkoreit, Llion Jones,
  Aidan~N. Gomez, Lukasz Kaiser, and Illia Polosukhin.
\newblock Attention is all you need, 2023.

\bibitem[Verma et~al.(2022)Verma, Fu, Yang, and Levine]{chai}
Siddharth Verma, Justin Fu, Mengjiao Yang, and Sergey Levine.
\newblock Chai: A chatbot ai for task-oriented dialogue with offline
  reinforcement learning, 2022.

\bibitem[Wang et~al.(2021)Wang, Novikov, Zolna, Springenberg, Reed, Shahriari,
  Siegel, Merel, Gulcehre, Heess, and de~Freitas]{wang2021critic}
Ziyu Wang, Alexander Novikov, Konrad Zolna, Jost~Tobias Springenberg, Scott
  Reed, Bobak Shahriari, Noah Siegel, Josh Merel, Caglar Gulcehre, Nicolas
  Heess, and Nando de~Freitas.
\newblock Critic regularized regression, 2021.

\bibitem[Xie et~al.(2024)Xie, Zhang, Chen, Li, Zhao, Cao, Hua, Cheng, Shin,
  Lei, et~al.]{xie2024osworld}
Tianbao Xie, Danyang Zhang, Jixuan Chen, Xiaochuan Li, Siheng Zhao, Ruisheng
  Cao, Toh~Jing Hua, Zhoujun Cheng, Dongchan Shin, Fangyu Lei, et~al.
\newblock Osworld: Benchmarking multimodal agents for open-ended tasks in real
  computer environments.
\newblock \emph{arXiv preprint arXiv:2404.07972}, 2024.

\bibitem[Yan et~al.(2023)Yan, Yang, Zhu, Lin, Li, Wang, Yang, Zhong, McAuley,
  Gao, Liu, and Wang]{yan2023gpt4v}
An~Yan, Zhengyuan Yang, Wanrong Zhu, Kevin Lin, Linjie Li, Jianfeng Wang,
  Jianwei Yang, Yiwu Zhong, Julian McAuley, Jianfeng Gao, Zicheng Liu, and
  Lijuan Wang.
\newblock Gpt-4v in wonderland: Large multimodal models for zero-shot
  smartphone gui navigation, 2023.

\bibitem[Yang et~al.(2023{\natexlab{a}})Yang, Prabhakar, Narasimhan, and
  Yao]{intercode}
John Yang, Akshara Prabhakar, Karthik Narasimhan, and Shunyu Yao.
\newblock Intercode: Standardizing and benchmarking interactive coding with
  execution feedback, 2023{\natexlab{a}}.

\bibitem[Yang et~al.(2023{\natexlab{b}})Yang, Liu, Han, Chen, Huang, Fu, and
  Yu]{yang2023appagent}
Zhao Yang, Jiaxuan Liu, Yucheng Han, Xin Chen, Zebiao Huang, Bin Fu, and Gang
  Yu.
\newblock Appagent: Multimodal agents as smartphone users.
\newblock \emph{arXiv preprint arXiv:2312.13771}, 2023{\natexlab{b}}.

\bibitem[Yao et~al.(2023)Yao, Chen, Yang, and Narasimhan]{webshop}
Shunyu Yao, Howard Chen, John Yang, and Karthik Narasimhan.
\newblock Webshop: Towards scalable real-world web interaction with grounded
  language agents, 2023.

\bibitem[Zeng et~al.(2023)Zeng, Liu, Lu, Wang, Liu, Dong, and
  Tang]{agenttuning}
Aohan Zeng, Mingdao Liu, Rui Lu, Bowen Wang, Xiao Liu, Yuxiao Dong, and Jie
  Tang.
\newblock Agenttuning: Enabling generalized agent abilities for llms, 2023.

\bibitem[Zhai et~al.(2024)Zhai, Bai, Lin, Pan, Tong, Zhou, Suhr, Xie, LeCun,
  Ma, and Levine]{zhai2024fine}
Yuexiang Zhai, Hao Bai, Zipeng Lin, Jiayi Pan, Shengbang Tong, Yifei Zhou,
  Alane Suhr, Saining Xie, Yann LeCun, Yi~Ma, and Sergey Levine.
\newblock Fine-tuning large vision-language models as decision-making agents
  via reinforcement learning.
\newblock \emph{arXiv preprint arXiv:2405.10292}, 2024.

\bibitem[Zhang et~al.(2024{\natexlab{a}})Zhang, Li, He, Zhang, Qiao, Qin, Ma,
  Kang, Lin, Rajmohan, et~al.]{zhang2024ufo}
Chaoyun Zhang, Liqun Li, Shilin He, Xu~Zhang, Bo~Qiao, Si~Qin, Minghua Ma,
  Yu~Kang, Qingwei Lin, Saravan Rajmohan, et~al.
\newblock Ufo: A ui-focused agent for windows os interaction.
\newblock \emph{arXiv preprint arXiv:2402.07939}, 2024{\natexlab{a}}.

\bibitem[Zhang et~al.(2024{\natexlab{b}})Zhang, Wu, Teng, Liao, Xu, Xiao, Wei,
  and Tang]{zhang2024android}
Jiwen Zhang, Jihao Wu, Yihua Teng, Minghui Liao, Nuo Xu, Xiao Xiao, Zhongyu
  Wei, and Duyu Tang.
\newblock Android in the zoo: Chain-of-action-thought for gui agents,
  2024{\natexlab{b}}.

\bibitem[Zhang and Zhang(2023)]{zhang2023look}
Zhuosheng Zhang and Aston Zhang.
\newblock You only look at screens: Multimodal chain-of-action agents, 2023.

\bibitem[Zhang et~al.(2024{\natexlab{c}})Zhang, Tian, Chen, and
  Liu]{zhang2024mmina}
Ziniu Zhang, Shulin Tian, Liangyu Chen, and Ziwei Liu.
\newblock Mmina: Benchmarking multihop multimodal internet agents.
\newblock \emph{arXiv preprint arXiv:2404.09992}, 2024{\natexlab{c}}.

\bibitem[Zheng et~al.(2024)Zheng, Gou, Kil, Sun, and Su]{zheng2024gpt4vision}
Boyuan Zheng, Boyu Gou, Jihyung Kil, Huan Sun, and Yu~Su.
\newblock Gpt-4v(ision) is a generalist web agent, if grounded, 2024.

\bibitem[Zhou et~al.(2023)Zhou, Xu, Zhu, Zhou, Lo, Sridhar, Cheng, Bisk, Fried,
  Alon, and Neubig]{Zhou2023WebArenaAR}
Shuyan Zhou, Frank~F. Xu, Hao Zhu, Xuhui Zhou, Robert Lo, Abishek Sridhar,
  Xianyi Cheng, Yonatan Bisk, Daniel Fried, Uri Alon, and Graham Neubig.
\newblock Webarena: A realistic web environment for building autonomous agents.
\newblock \emph{ArXiv}, abs/2307.13854, 2023.
\newblock URL \url{https://api.semanticscholar.org/CorpusID:260164780}.

\bibitem[Zhou et~al.(2024)Zhou, Zanette, Pan, Levine, and
  Kumar]{zhou2024archer}
Yifei Zhou, Andrea Zanette, Jiayi Pan, Sergey Levine, and Aviral Kumar.
\newblock Archer: Training language model agents via hierarchical multi-turn
  rl.
\newblock \emph{arXiv preprint arXiv:2402.19446}, 2024.

\bibitem[Ziegler et~al.(2019)Ziegler, Stiennon, Wu, Brown, Radford, Amodei,
  Christiano, and Irving]{ziegler2019finetuning}
Daniel~M. Ziegler, Nisan Stiennon, Jeffrey Wu, Tom~B. Brown, Alec Radford,
  Dario Amodei, Paul~F. Christiano, and Geoffrey Irving.
\newblock Fine-tuning language models from human preferences.
\newblock \emph{CoRR}, abs/1909.08593, 2019.
\newblock URL \url{http://arxiv.org/abs/1909.08593}.

\end{thebibliography}
\bibliographystyle{plainnat}
\newpage
\appendix



\part*{Appendices}
\section{Environment details}

\subsection{Post-processing of AitW}
\label{app:env_details}
The Android in the Wild (AiTW) task set is a large-scale dataset for android device control, containing five subsets: GoogleApps, Install, Web Shopping, General, and Single, where we select the General and Web Shopping subsets. Single subset is not considered here because all tasks in Single can be completed within one step and thus this subset fails to examine the multi-step challenges that we are interested in this paper. Install and GoogleApps are not considered due to security reasons as those tasks require an active Google account and parallel emulations can flag security concerns.

\textbf{General.} The General set focuses on searching for information and basic application usage. For example, it contains searching for latest news in Chile,  search for flights from NYC to Sydney, opening Gmail, etc. We use all 545 tasks in the training set for training and the first 96 tasks in the test set for testing due to computational and budget constraints. The maximum allowed number of steps for this subset is 10. Offline data is collected by rolling our the initial AutoUI policy with tasks from the training set. The offline data used for the offline-to-online setting contains 608 trajectories while the offline data used for the offline setting contains 1552 trajectories. Some task examples are shown in~\Cref{tab:webshop-curriculum}.

\begin{table}[htbp]
    \centering
    \begin{tabular}{@{}l@{}}
        \toprule
        Task Example \\
        \midrule
        How do I get to the nearest Verizon Store?\\
        How much does a 2 bedroom apartment rent for in Denver?\\
        Search for flights from Barcelona to Boston\\
        What's a good restaurant in New York?\\
        What's on the menu at Burger King?\\
        \bottomrule
    \end{tabular}
    \caption{Examples of task descriptions in the AiTW General task set.}
    \label{tab:general-tasks}
\end{table}

\textbf{Web Shopping.} The Web Shopping subset comprises search instructions on various shopping websites, like searching for razer blader on ebay. As some websites (e.g. Amazon) and operations (e.g. adding items to cart) frequently require captcha verifications, we post-process the Web Shopping subset to exclude such operations and websites and also make the task easy to evaluate for our autonomous evaluator. The resulting task set involves navigating through five websites (costco.com, bestbuy.com, target.com, walmart.com, newegg.com) and three basic operations (go to website, search in the website, and select items from the searched results). Our post-processed training set contains 438 tasks and our testing set contains 96 tasks. Example tasks after post-processing can be found in Table~\ref{tab:webshop-curriculum}. The maximum allowed number of steps for this subset is 20. Offline data is collected by rolling our the initial AutoUI policy with tasks from the training set. The offline data used for the offline-to-online setting contains 528 trajectories while the offline data used for the offline setting contains 1296 trajectories. 


\begin{table}[ht]
    \centering
    \begin{tabular}{@{}cl@{}}
        \toprule
        Difficulty & Task Example \\
        \midrule
        \multirow{2}{*}{1} & Go to costco.com \\
                           & Go to walmart.com \\
        \midrule
        \multirow{2}{*}{2} & Go to costco.com, search for "bose soundsport free" \\
                           & Go to walmart.com, search for "logitech g910" \\
        \midrule
        \multirow{2}{*}{3} & Go to costco.com, search for "bose soundsport free" and select the first entry\\
                           & Go to walmart.com, search for "logitech g910" and select the first entry\\
        \bottomrule
    \end{tabular}
    \caption{Examples of task descriptions in the AiTW Webshopping task set.}
    \label{tab:webshop-curriculum}
\end{table}

\section{Other Quantitative Experiments}

\subsection{Horizon Limit}

\begin{figure}[!t]
     \centering
     \begin{subfigure}[b]{0.46\textwidth}
         \centering
    \includegraphics[width=\textwidth]{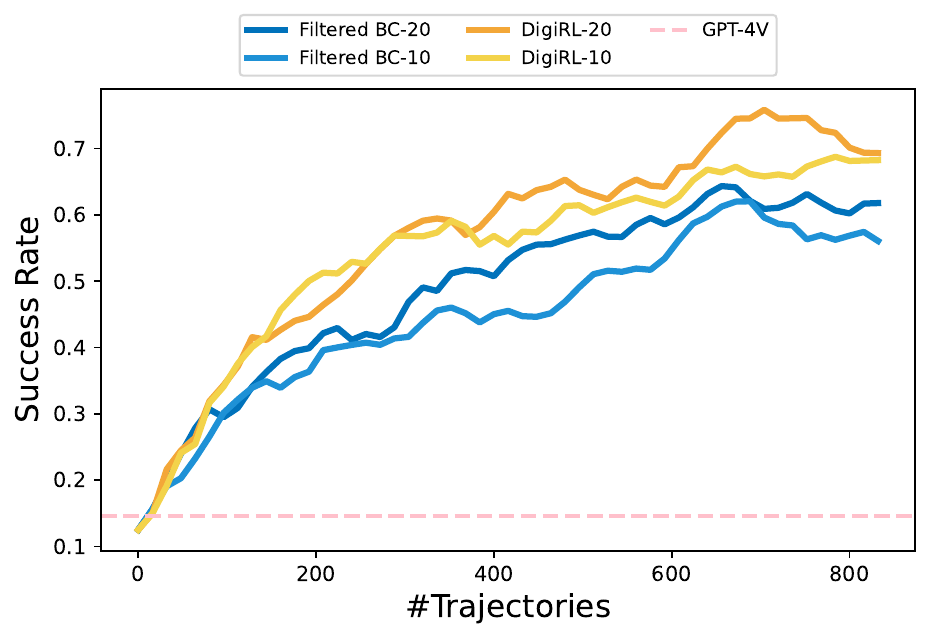}
     \end{subfigure}
    \caption{\textbf{Success rate with different horizon length} ($H\in\{10, 20\}$)under different methods on the AiTW Google Search task set.}
    \label{fig:curriculum-and-maxstep}
\end{figure}

We investigate the horizon limit of filtered BC and DigiRL on the AitW General subset. As most tasks can be effectively solved within 10 steps, we specify two horizon limits: a sufficient horizont $H=10$, and a redundant horizon $H=20$. Results in~\Cref{fig:curriculum-and-maxstep} show that a redundant horizon introduces significantly faster learning speed for both filtered BC and DigiRL, presumbaly because longer horizon means more opportunity to try in a single trajectory. In both horizon settings, we observe the \ouragent offers a significant speedup of around 100 trajectories over Filtered BC.

\subsection{Trajectory Length}

\begin{table}[!t]
    \centering
    \small
    \setlength{\tabcolsep}{5.0pt}
        \begin{tabular}{ccccc}
            \toprule
            &\multicolumn{2}{c}{\textbf{AitW General}} & \multicolumn{2}{c}{\textbf{AitW Web Shopping}} \\ 
            \cmidrule(lr){2-3} \cmidrule(lr){4-5}
             & All Trajectories & Successful Trajectories & All Trajectories & Successful Trajectories \\
            \midrule
            DigiRL Run1 & \gcr{6.31} & \gct{4.40} & \gcs{11.35} & \gcu{7.23} \\
            DigiRL Run2 & \gcr{6.64} & \gct{5.04} & \gcs{10.86} & \gcu{6.55} \\
            Filtered BC Run1 & \gcr{8.08} & \gct{6.56} & \gcs{12.05} & \gcu{6.88} \\
            Filtered BC Run2 & \gcr{7.36} & \gct{6.13} & \gcs{14.72} & \gcu{9.62} \\
            \bottomrule
        \end{tabular}
    \caption{\footnotesize{\textbf{Average rollout length of the DigiRL agent compared to filtered BC.} Darker green means shorter rollout length. On both AitW General and AitW Web Shopping test subsets, we find that DigiRL consistently produces shorter length rollouts than filtered BC.}}
    \label{tab:traj-lengths}
    \vspace{-0.35cm}
\end{table}

We investigate the rollout length of DigiRL compared to filtered BC. Results in~\Cref{tab:traj-lengths} demonstrate that DigiRL consistently achieves shorter average rollout lengths compared to filtered BC across both subsets. This observation holds true whether considering all rollouts for computing this correlation or only investigating this correlation on rollouts that eventually succeed. This indicates the capability of DigiRL to solve tasks in a more efficient and directed manner. Qualitative examples can be found in~\Cref{fig:traj-len-example}.

\section{Qualitative Examples}
\subsection{Random sample of trajectories for different agents}
\begin{figure}
    \centering
    \begin{subfigure}{\textwidth}
        \centering
        \includegraphics[width=0.75\textwidth]{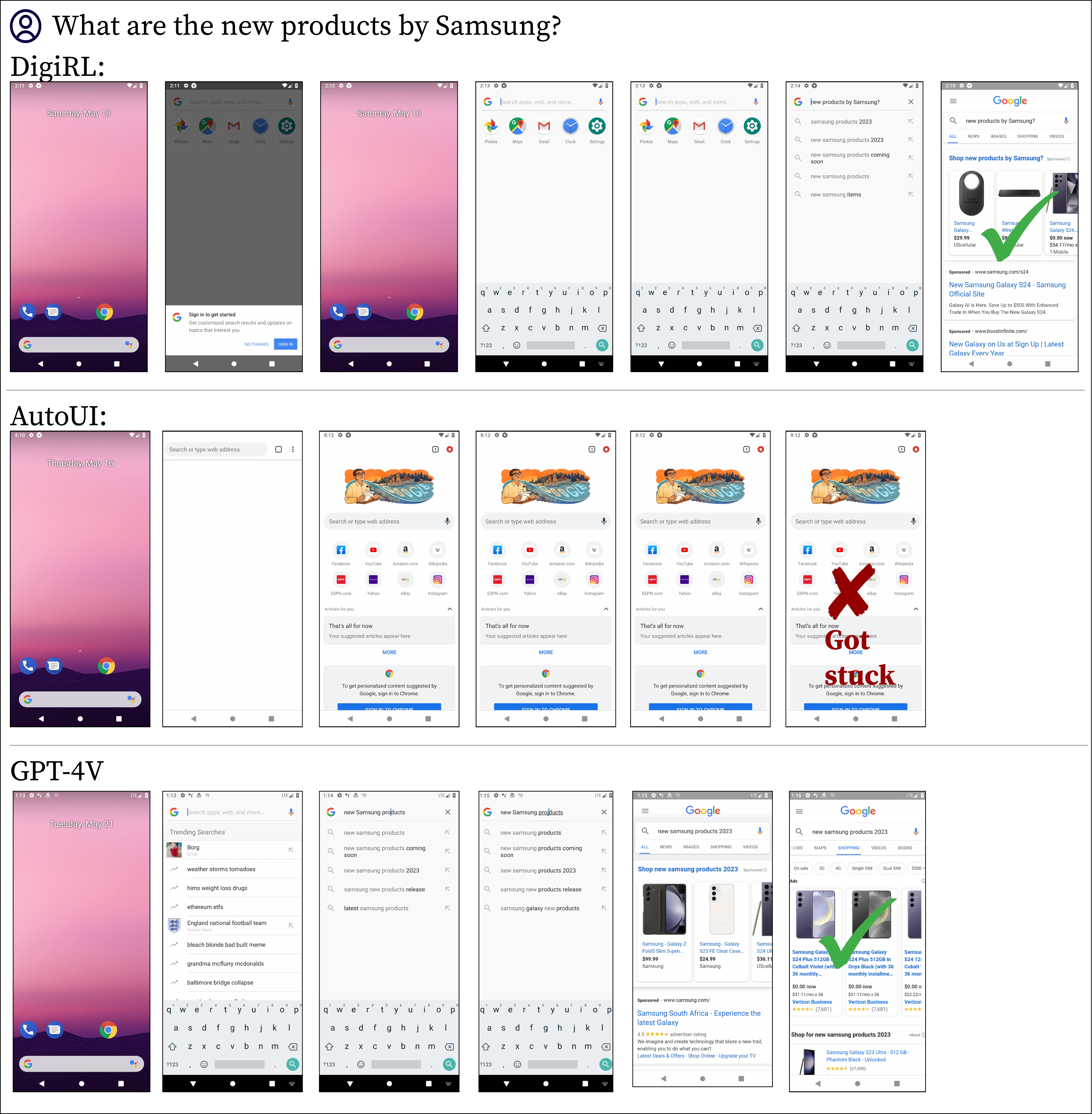}
    \end{subfigure}
    \begin{subfigure}{\textwidth}
        \centering
        \includegraphics[width=0.75\textwidth]{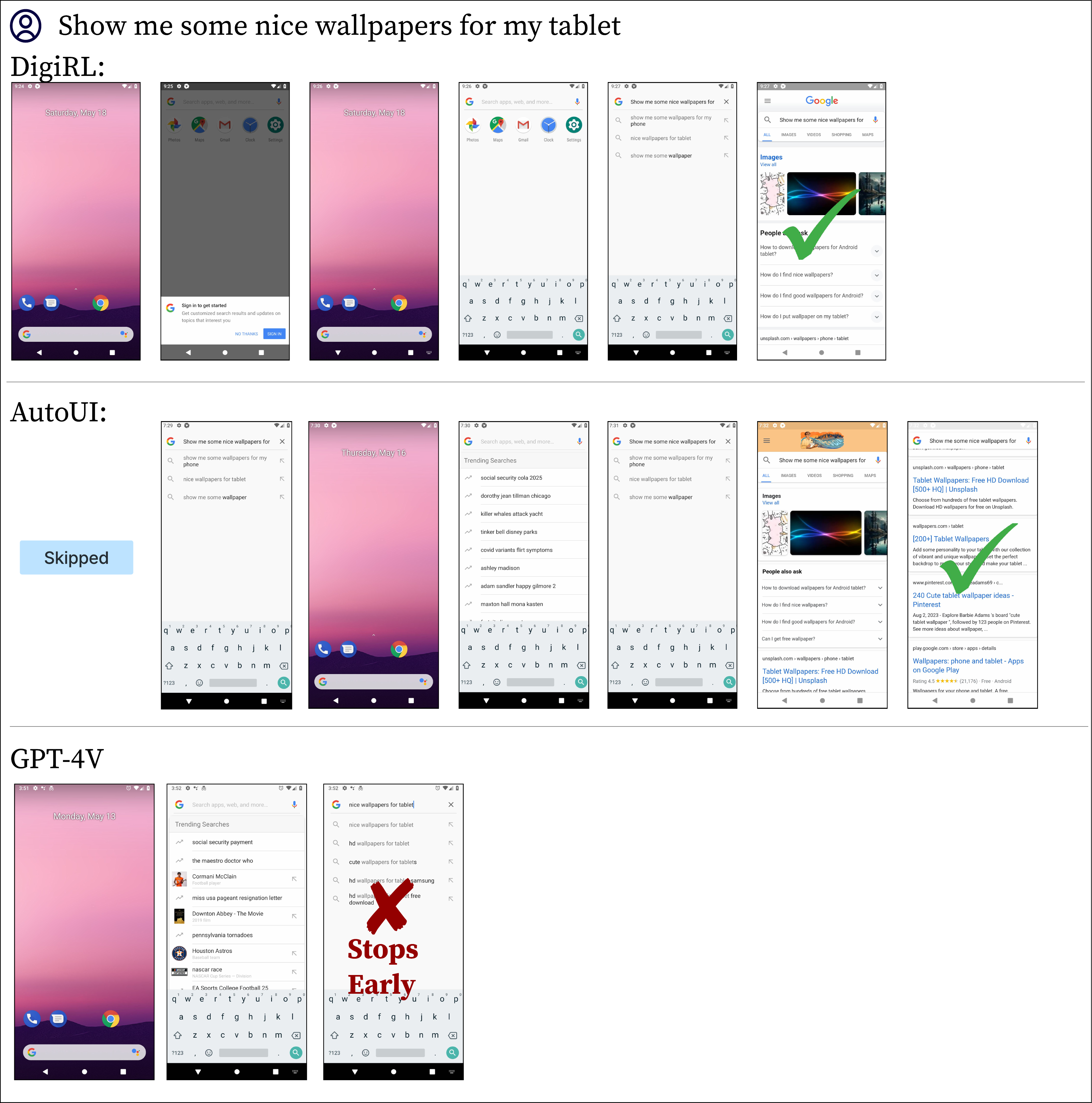}
    \end{subfigure}
    \caption{Agents' trajectory on two randomly sampled tasks on the General split of AitW.}
    \label{fig:example-general}
\end{figure}

\begin{figure}
    \centering
    \begin{subfigure}[b]{\textwidth}
        \centering
        \includegraphics[width=0.75\textwidth]{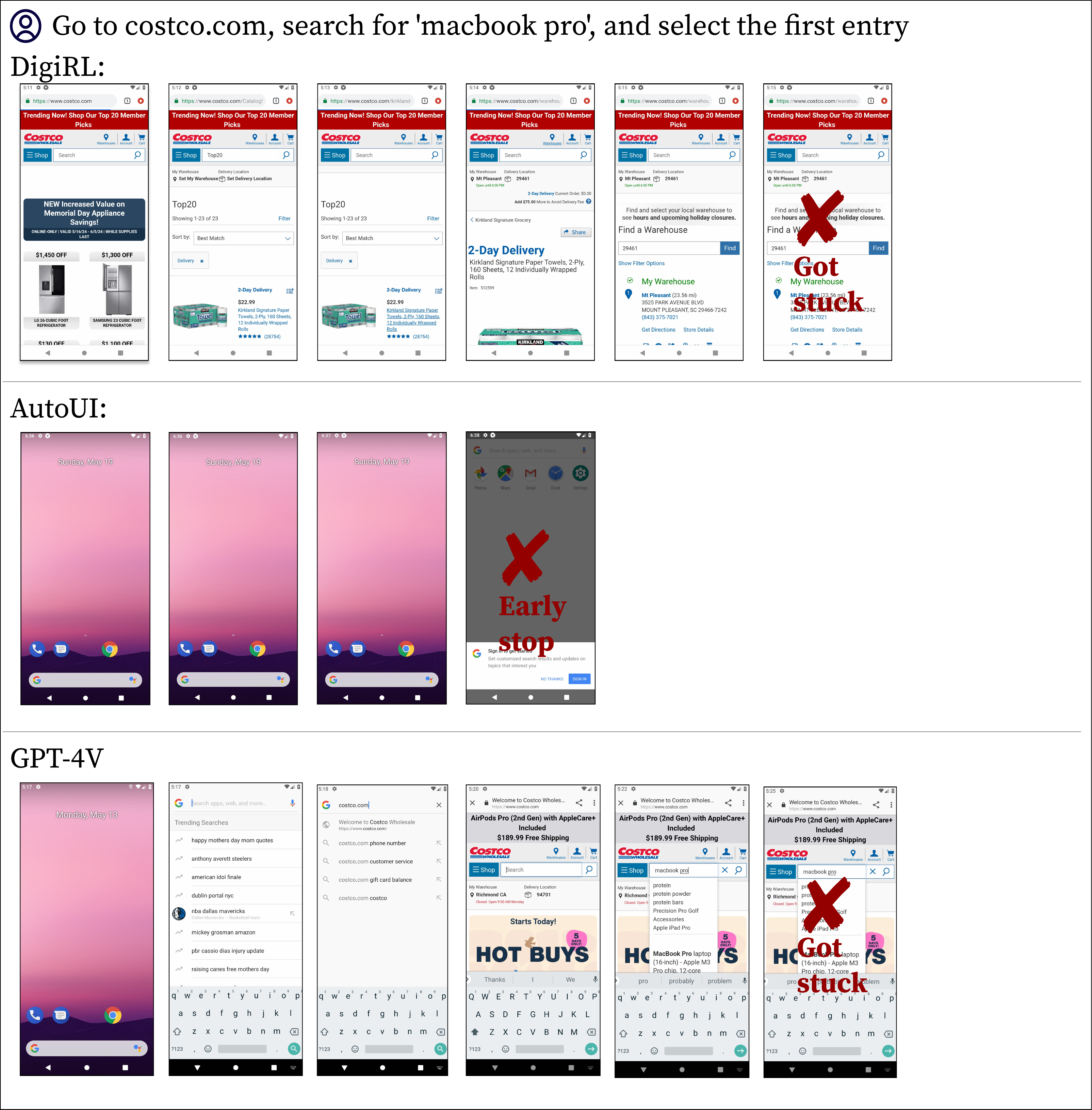}
    \end{subfigure}
    \begin{subfigure}[b]{\textwidth}
        \centering
        \includegraphics[width=0.75\textwidth]{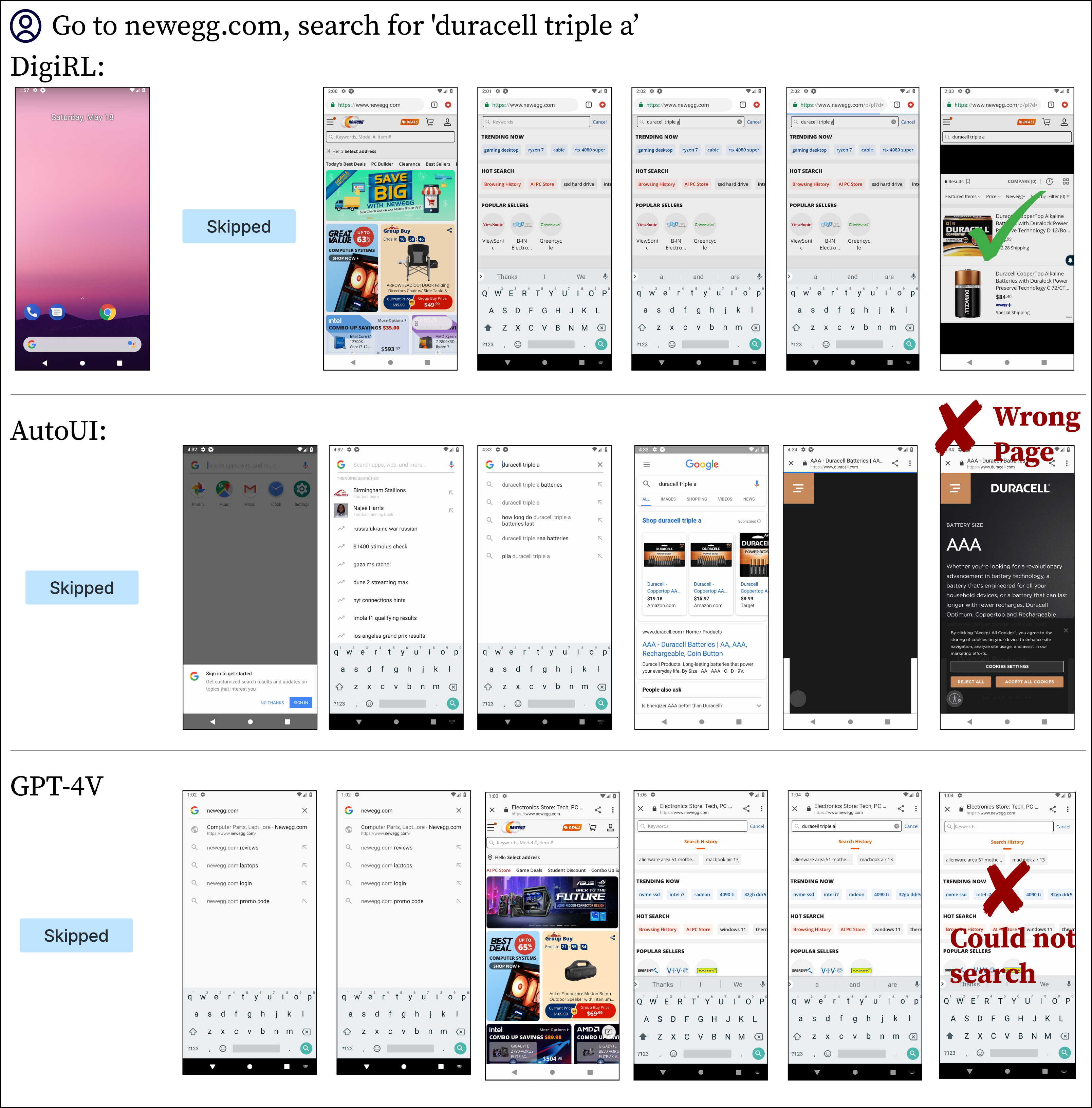}
    \end{subfigure}
    \caption{Agents' trajectory on two randomly sampled tasks on the WebShop split of AitW.}
    \label{fig:example-shop}
\end{figure}

In Figures~\ref{fig:example-general} and~\ref{fig:example-shop}, we provide trajectories of \ouragentnospace, AutoUI, and GPT-4V randomly sampled from our test set to offer a qualitative understanding of the agents’ performance. As shown in these examples, \ouragentnospace can efficiently carry out in-the-wild device control tasks and less likely to get stuck or get to a wrong page compared to AutoUI and GPT-4V.

\subsection{Error Recovery} \label{app:error-recovery}

\begin{figure}
    \centering
    \includegraphics[width=0.95\textwidth]{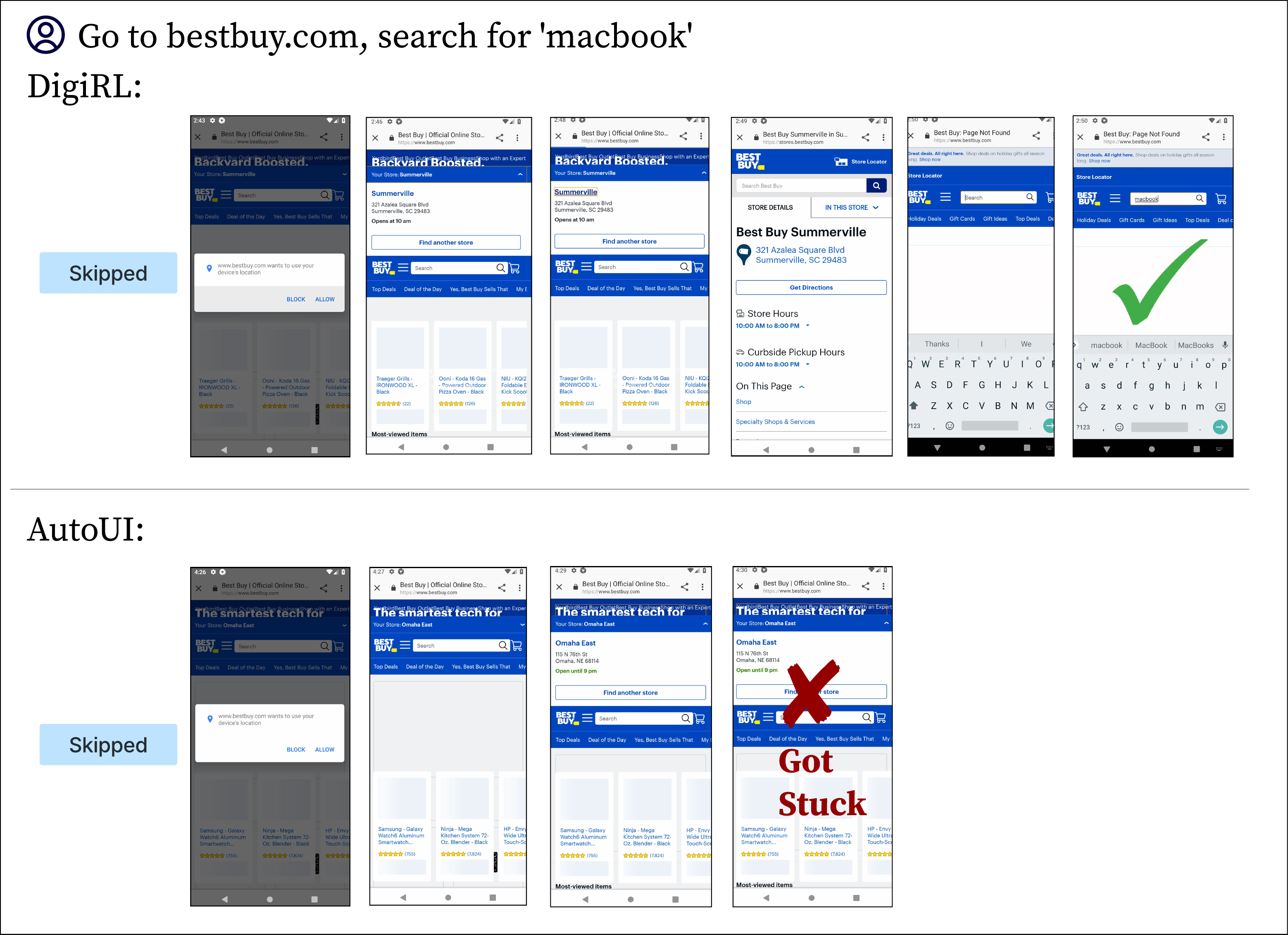}
    \caption{Error recovery cases. In \texttt{bestbuy.com}, we systematically find DigiRL able to recover from its own mistakes, while AutoUI fails to do so.}
    \label{fig:recover-from-mistakes}
\end{figure}

We observe that DigiRL is able to recover from its own mistakes. As shown in~\Cref{fig:recover-from-mistakes}, we find that DigiRL explores ways to get back to the original screen in order to perform a search. As a comparison, AutoUI fails to reset to the original screen and gets stuck at the diverged screen. Under the hood, we find DigiRL trying to maximize the state value, which usually induces it to reset to the original screen (that has a large value to success).

\subsection{Trajectory Length} \label{app:traj-length}

\begin{figure}[t]
    \centering
    \includegraphics[width=0.95\textwidth]{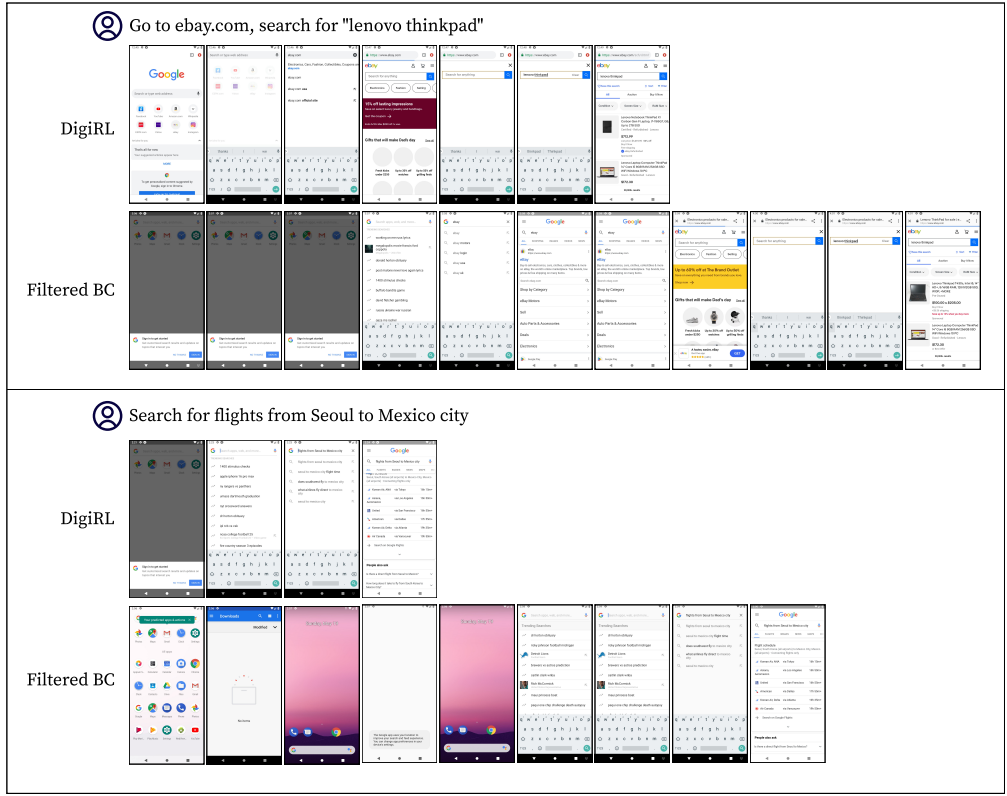}
    \caption{Examples where DigiRL has shorter trajectory length than online filtered BC.}
    \label{fig:traj-len-example}
\end{figure}

Qualitative example on the number of steps in trajectories of DigiRL and filtered BC are shown in~\Cref{fig:traj-len-example}. We find consistent cases where DigiRL has shorter trajectory length than filtere BC.

\subsection{Reasoning failure of GPT-4V}
\begin{figure}
    \centering
    \includegraphics[width=1.0\textwidth]{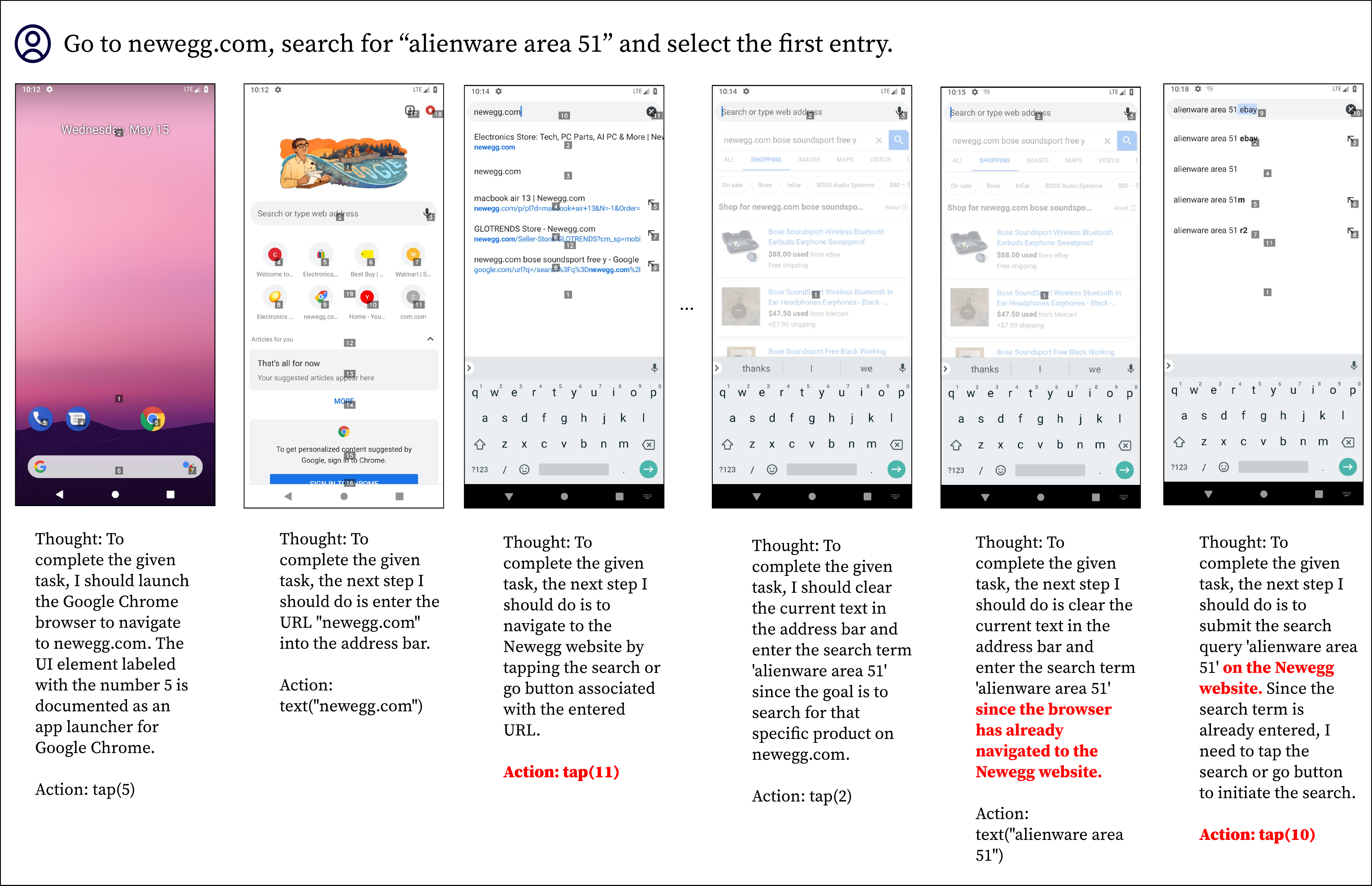}
    \caption{Failure of GPT-4V, with its thoughts and link-based actions given. A typical cause of failure is that it cannot tap on the correct ``search'' button after entering a query and mistakenly tapped onto the ``x'' symbol in the search bar as the ``search'' button. Here the goal is: Go to newegg.com, search for ``alienware area 51'' and select the first entry. As seen in red emboldened actions, it fails to press search button and deletes the query instead. Also, as seen in red highlighted parts in thoughts, it thinks it is in ``newegg.com'' website even though it is not.}
    \label{fig:gpt4_search_failure}
\end{figure}

\begin{figure}
    \centering
    \includegraphics[width=\textwidth]{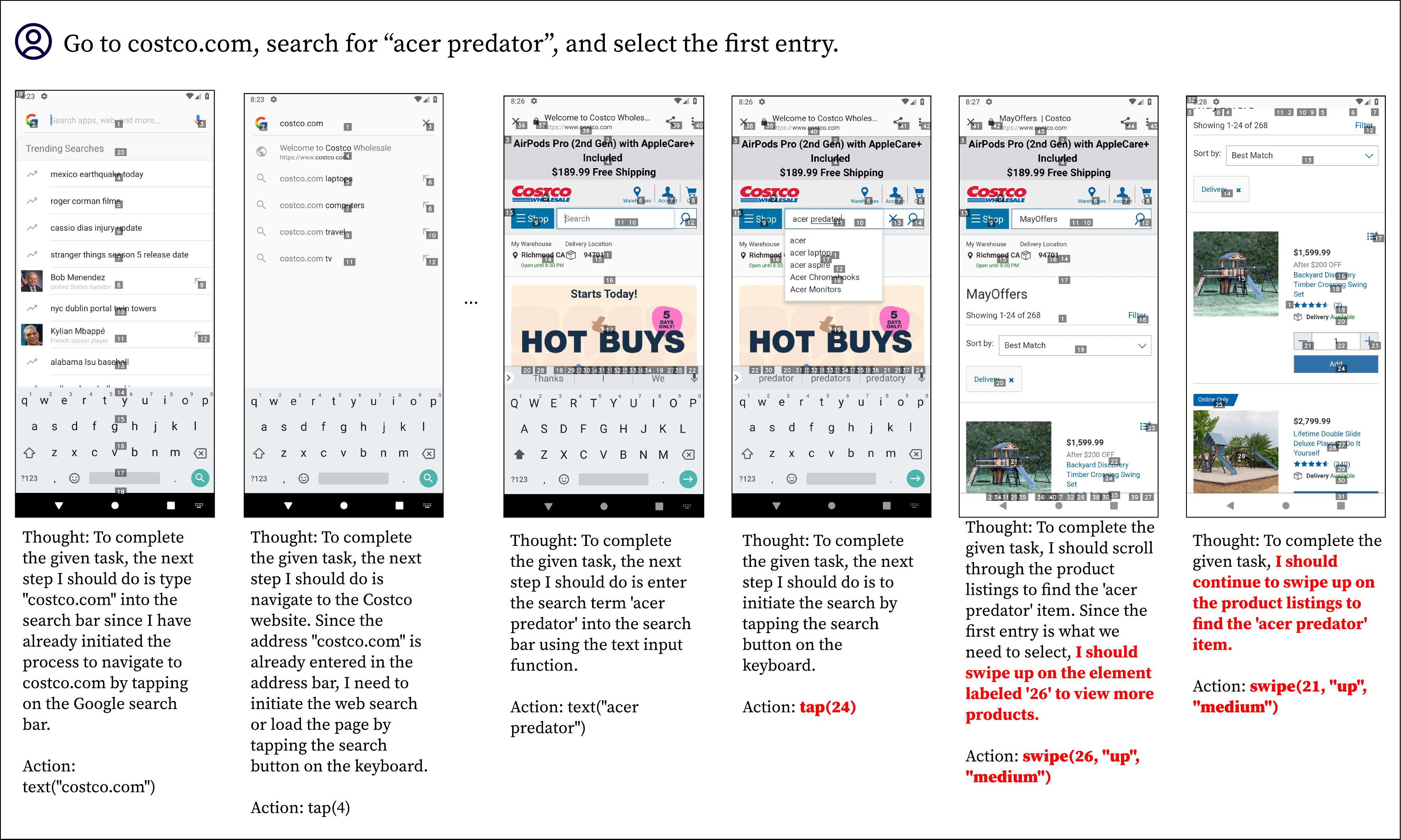}
    \caption{Failure of GPT-4V, with its thoughts and link-based actions given. This time the reason for failure is misclick on the wrong button. The task is ``Go to costco.com, search for ``acer predator'', and select the first entry''. Notice that up until the fourth frame in this Figure, the trajectory goes correct. But then it clicks on the generic advertisements on the Costco.com website, and it cannot recover back. It continues to scroll the page and takes wrong actions thereafter.}
    \label{fig:gpt4_recovery_failure}
\end{figure}

\begin{figure}
    \centering
    \includegraphics[width=1.02\textwidth]{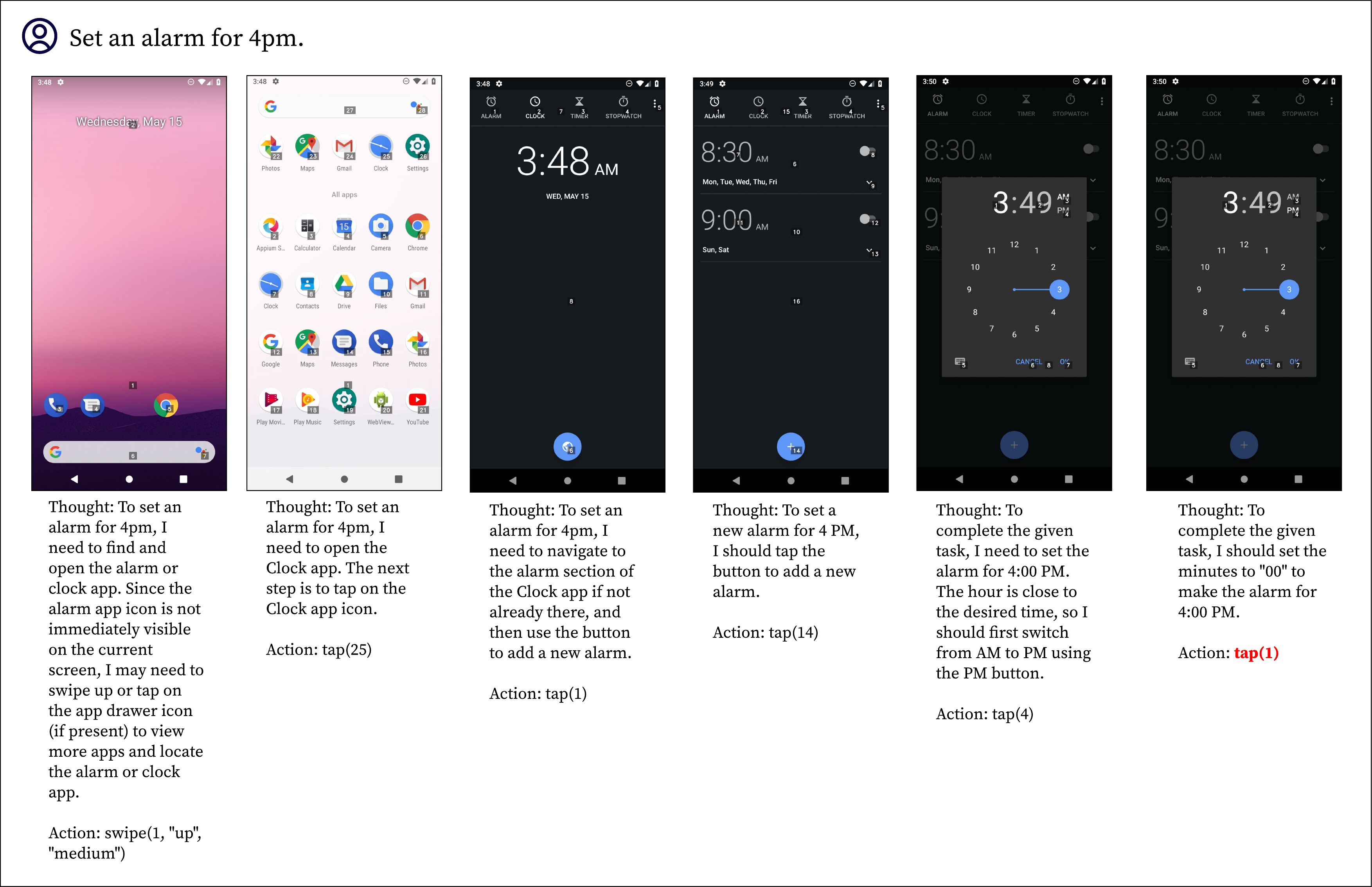}
    \caption{Failure of GPT-4V, with an example task on the AiTW general test set. The task is ``Set an alarm for 4pm''. Here, GPT-4V is able to successfully navigate to the clock app, and the alarm settings of that app. However, it cannot take the correct precise actions to set the alarm quickly enough, and it fails due to maximum rounds reached. In the last round, notice that the action of tap(1) contradict with its own thought process of setting minutes to ``00''.}
    \label{fig:gpt4_clock_failure}
\end{figure}

The performance of GPT-4V failed on AiTW tasks predominantly due to not being able to carry out control actions as it plans on a high level, and then not being able to recover from these mistakes. Moreover, one of the main reasons why it is not able to recover from a mistake is that it might hallucinate and make itself believe that it is a wrong app or website. Indeed, GPT-4V constructs a plan of further actions when provided a task from either Web Shopping or General dataset of AiTW. Then, when it makes a misclick and fails to successfully proceed in an intermediate step, it might think that it actually solved that intermediate step and is in the correct app or website to execute further actions, causing the overall trajectory to fail. An example of this is provided in Figure \ref{fig:gpt4_search_failure}. Here, we ask the model to search for an item in a webshopping website, in particular in ``newegg.com''. However, the model fails to proceed to that website due to not being able to precisely locating the search button. Then, instead of trying to go to that website again, the model thinks it is already in that webshopping website, and mistakes the search bar of Google with the search bar of ``newegg.com''. Hence, the rest of the trajectory also fails. Another slightly different phenomenon is illustrated in Figure \ref{fig:gpt4_recovery_failure}. Here, the model is able to proceed to the correct website and search for an item, but this time it fails to tap on the search button on the website and clicks to an advertisement instead. Consequently, the model fools itself to think it successfully searched the item, and scrolls the page hoping to find that item, but it cannot do so because in reality it views the results of the advertisement. The primary reason of these failures is the challenge of grounding the control actions in GUI interfaces to realize the intermediary goals laid out by GPT-4V model's thoughts. As an example, we provide an illustration of trying to set up an alarm task in Figure \ref{fig:gpt4_clock_failure}. Here, in the last frame, it fails to execute the precise movements in the necessary amount of rounds to correctly set up the alarm to the desired time, and in the last frame we see that the action taken does not align with the thought process of the model.

\section{Fine-grained failure modes}\label{app:fine-grained}
\begin{figure}
     \centering
     \begin{subfigure}[b]{\textwidth}
         \centering
    \includegraphics[width=\textwidth]{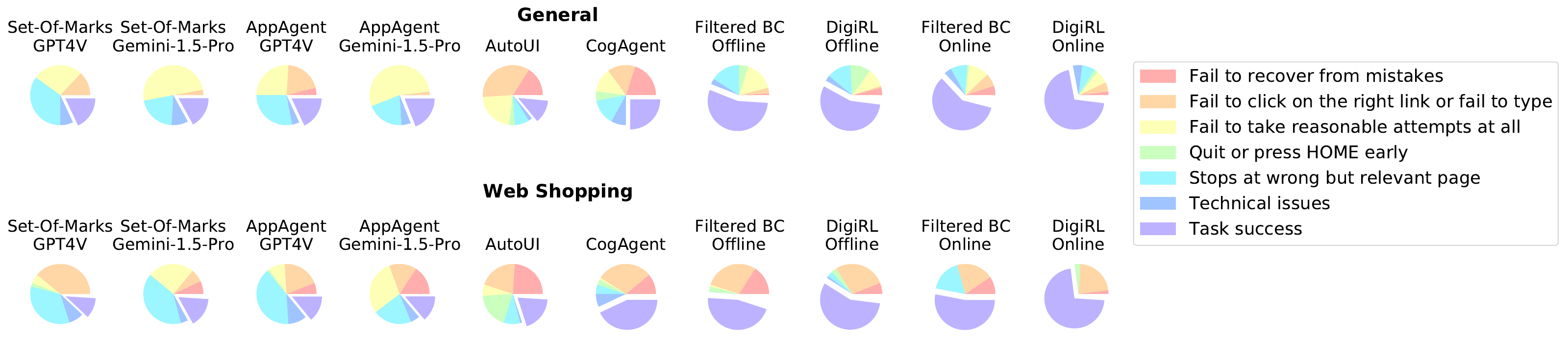}
     \end{subfigure}
    \caption{Failure modes decomposition for each policy model for both General and Web Shopping subsets.}
    \label{fig:fine_grained}
\end{figure}

In Figure~\ref{fig:fine_grained}, we present a more fine-grained breakdown for all six failure modes provided in the user study. Those failure modes include:
\begin{itemize}
    \item \textit{Failure to recover from mistakes} refers to the scenario where the agent made a mistake that led it to states from which it failed to quickly recover and resume the task, such as a wrong google search page.
    \item \textit{Failure to click on the right link or failure to click} refers to the failure mode where the agent either fails to locate the element that it tries to click on and keeps clicking on the nearby region, or fails to start typing in the string when it is supposed to do so.
    \item \textit{Failure to take reasonable attempts at all} refers to the failure mode where there is no clear reason that the agent fails to complete the task and does not seem to be on the right track throughout the trajectory.
    \item \textit{Quit or press HOME early} refers to the failure mode where the agent decided to finish the task or press HOME to start over before the task is actually finished.
    \item \textit{Stops at wrong but relevant page} refers to the failure mode where the agent arrives at a wrong page and mistakenly thinks that it had completed the task. For example, the agent finds a macbook on costco.com while the instruction asked it to find a macbook on ebay.com.
    \item \textit{Technical issues} refer to the failure mode that either the task is impossible (e.g. the tasks asks to open Amazon app but this app is not installed) or the agent is temporarily blocked from a certain website due to frequent visits.
\end{itemize}
The translation between fine-grained failure modes and coarse-grained failure modes is presented in Table~\ref{tab:translation}. 
\begin{table}[htbp]
    \centering
    \begin{tabular}{@{}cl@{}}
        \toprule
        Fine-Grained Failure & Coarse-Grained Failure \\
        \midrule
        Fail to recover from mistakes &  Fail to recover from mistakes\\
        Fail to click on the right link or fail to type &  Get stuck midway\\
        Fail to take reasonable attempts at all&  Get stuck midway\\
        Quit or Press HOME early &  Arrive at wrong goal\\
        Stops at wrong but relevant page&  Arrive at wrong goal\\
        Technical Issues&  None\\
        \midrule
    \end{tabular}
    \caption{Examples of task descriptions in the AiTW Webshopping task set.}
    \label{tab:translation}
\end{table}






\section{Experiment machines} \label{app:machines}
Our main experiments are conducted on VM instances from Google Cloud Platform. Each VM instance comes with 1x Tesla T4 GPU and 16x Intel(R) Xeon(R) CPU.
\section{Setup for parallel environment} \label{app:parallel}
Running multiple emulators in parallel can be challenging due to the inefficiency in thread synchronization and frequent fault propagation when one emulator runs into an unknown error. To address this challenge, we set up a server-client system where all emulator processes are running in independent server processes. Each emulator process communicates with the main training process through different UIAutomotor servers. The main training process sends high-level instructions to UIAutomotor servers (such as reset and step), while UIAutomotor servers parse high-level instructions into low-level UI commands (such as typing a character and tapping at a coordinate) and such UI commands are executed by the emulator processes. When an exception is thrown in the emulator, the UIAutomotor examines if it is recoverable (e.g. an UI command takes too long to execute in the emulator) and reset the emulator process if it is not. When an exception is thrown in the UIAutomotor server, the main training process stops and resets the UIAutomotor server to ensure data correctness. 

\begin{figure}[t]
     \centering
     \begin{subfigure}[b]{0.96\textwidth}
         \centering
    \includegraphics[width=\textwidth]{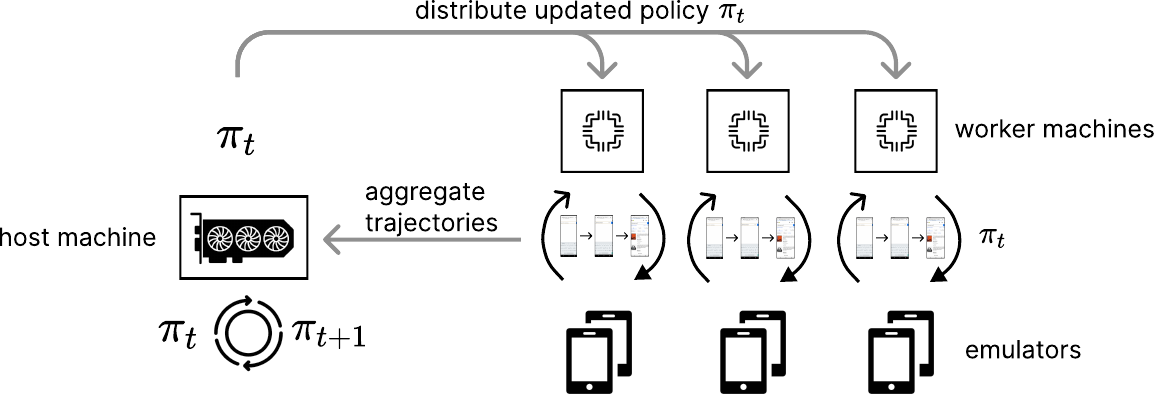}
     \end{subfigure}
    \caption{Multi-machine parallel emulator execution. The host machine is equipped with GPU accelerators and the worker machines are equipped only with CPUs. The policy update is executed on the worker machine and the trajectory collections are executed distributedly on the worker machines and aggregated by the host machine.}
    \label{fig:distr-arch}
\end{figure}

This design can easily be scaled up to a multi-machine setting. As illustrated in~\Cref{fig:distr-arch}, one host machine equipped with GPU accelerator has a local copy of the current policy $\pi_t$, and distributes the policy to all worker machines equipped with only one GPU and multiple CPUs. Each worker machine will then collect trajectories of different tasks using $\pi_t$. After all collection processes are synchronized, the host machine gathers all the trajectories together to update the policy to $\pi_{t+1}$. This process keeps iterating until the policy converges.


\section{Autonomous evaluator details}

Our autonomous evaluator gives a reward to each observation we get. The observation is composed of the current screenshot of device and the task. The evaluator gives a reward of $1$ if the screenshot shows a completion of the task, and will terminate the POMDP as a result result.

\begin{figure}[htbp]
     \centering
     \begin{subfigure}[b]{0.98\textwidth}
         \centering
    \includegraphics[width=\textwidth]{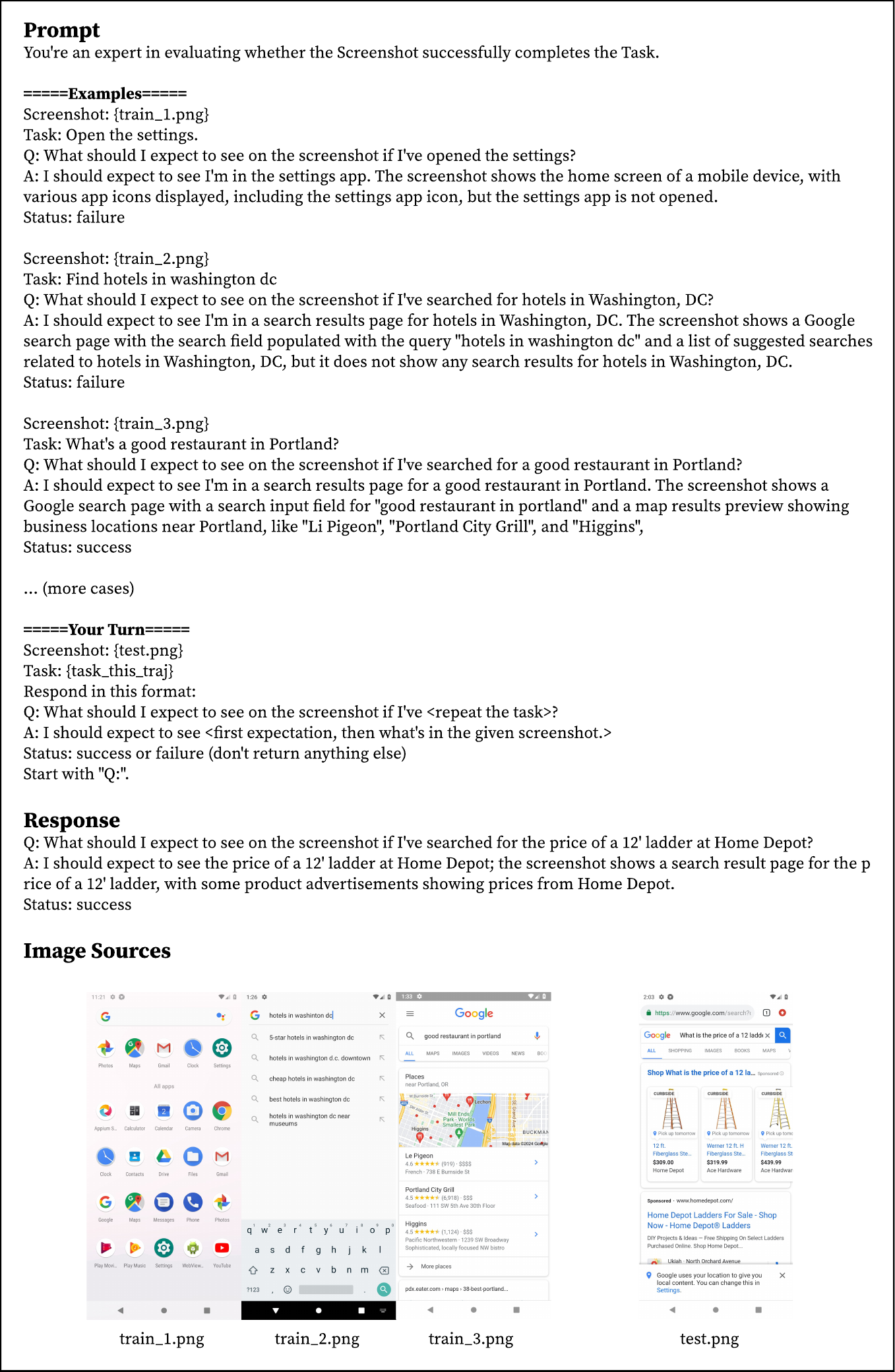}
     \end{subfigure}
    \caption{Prompt for our autonomous evaluator for tasks in AitW General subset.}
    \label{fig:eval-prompt-1}
\end{figure}

\begin{figure}[htbp]
     \centering
     \begin{subfigure}[b]{0.98\textwidth}
         \centering
    \includegraphics[width=\textwidth]{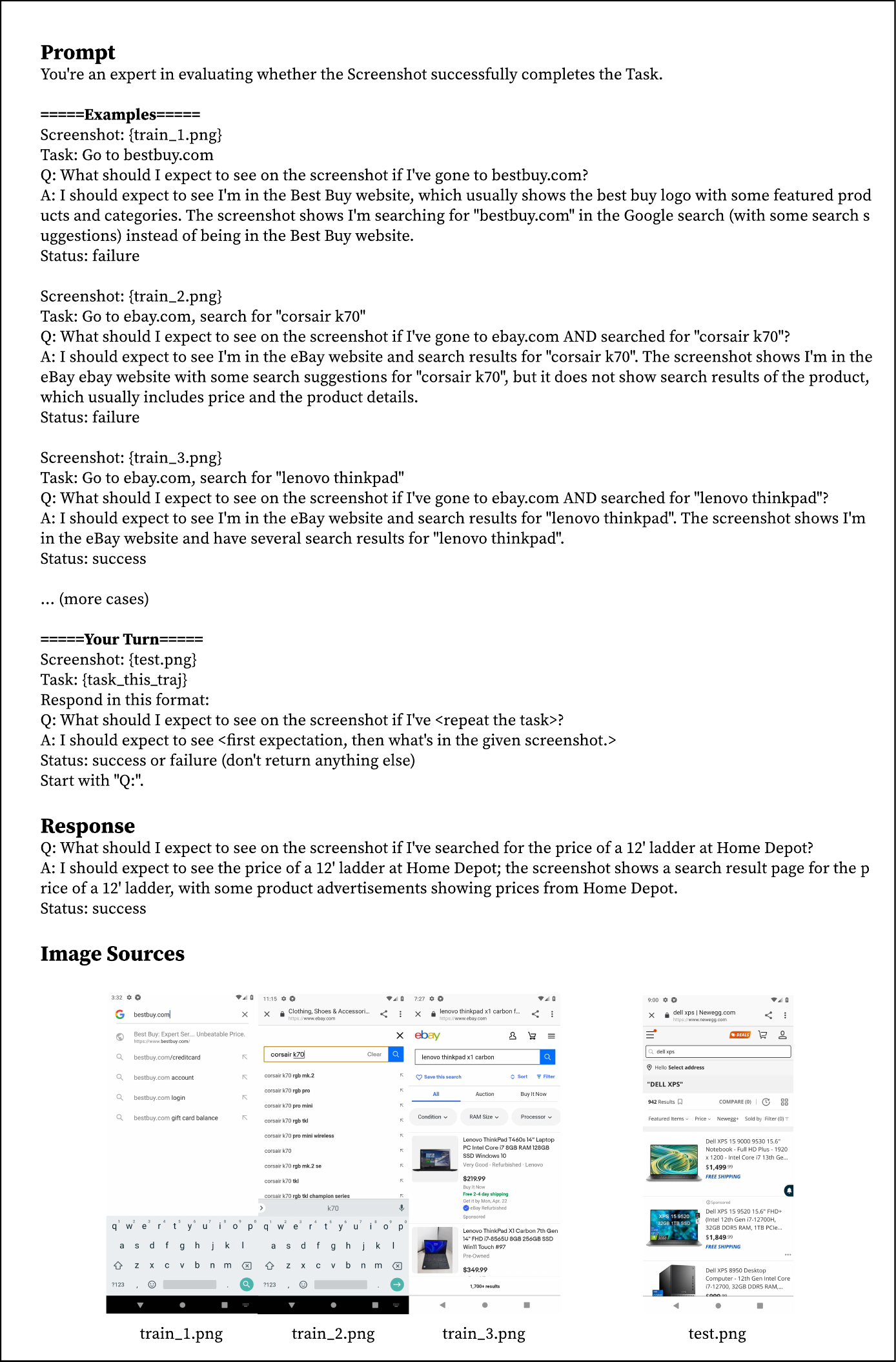}
     \end{subfigure}
    \caption{Prompt for our autonomous evaluator for tasks in AitW Web Shopping subset.}
    \label{fig:eval-prompt-2}
\end{figure}

The optimized prompt is shown in~\Cref{fig:eval-prompt-1} and ~\Cref{fig:eval-prompt-2} for General and Web Shopping subsets respectively.

\section{Zero-shot Baseline Details}
\begin{figure}[ht]
     \centering
    \includegraphics[width=\textwidth]{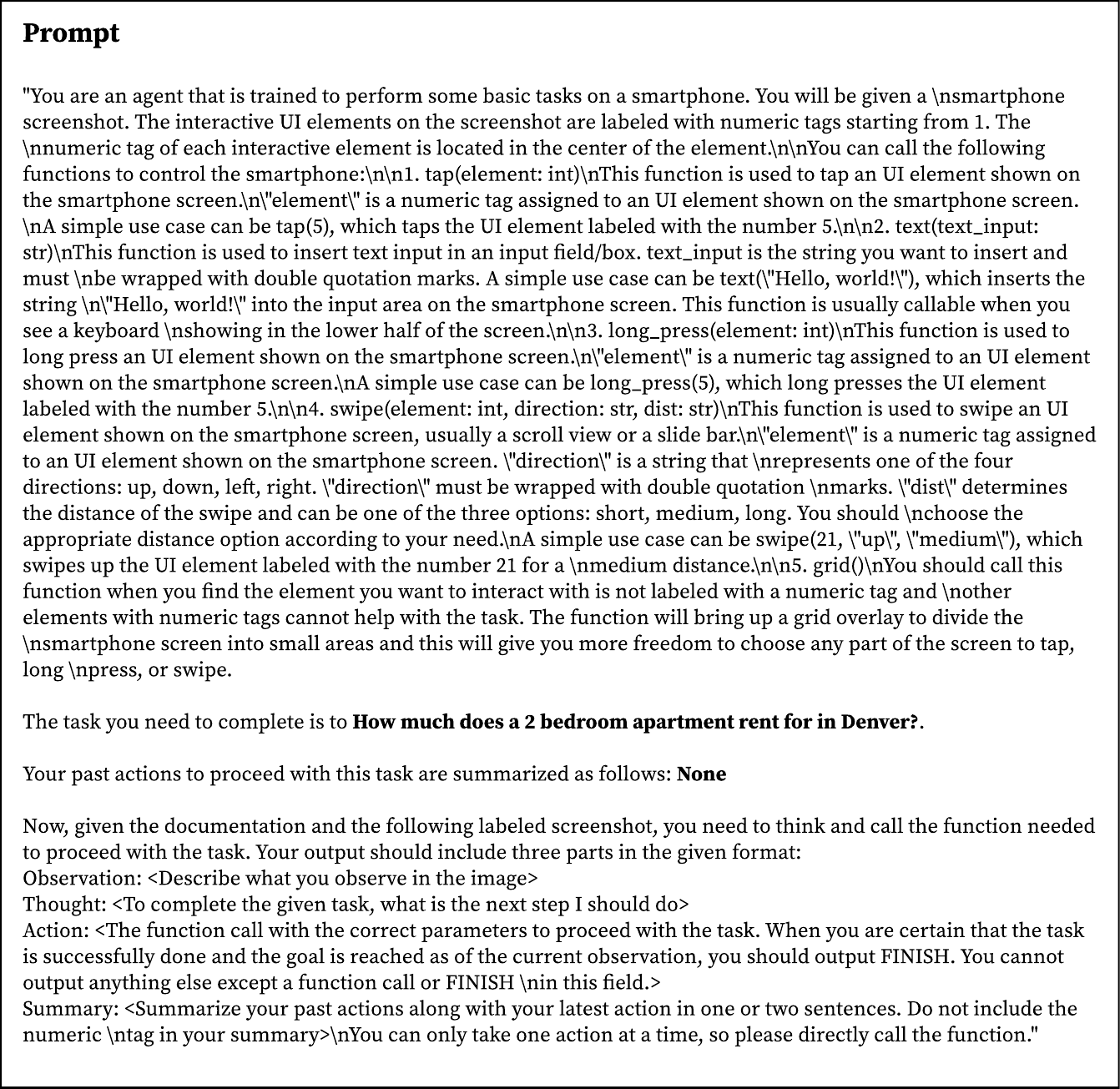}
    \caption{Set-of-Marks prompting. The boldened inputs can be changed according to our goal. The task changes for every different task. The past actions change as we take actions (it is None now since this is the prompt for the first round).}
    \label{fig:gpt4v-prompt}
\end{figure}

Figure~\ref{fig:gpt4v-prompt} shows the prompt that we used for testing the Set-of-Marks performance for GPT-4V and Gemini 1.5 Pro. This prompt is directly taken from \citet{yang2023appagent}.





\section{Hyperparameters} \label{app:hyperparameters}
Hyperparameters for both Filtered BC and \ouragent are carefully tuned through binary search on the training set of General and Web Shopping subsets. The final choice of hyperparameters for both methods can be found in Table~\ref{tab:hyperparameters}. As shown in the table, the only hyperparameters introduced by \ouragent are supervised training hyperparameters for the value function and instruction value function (including number of iterations and learning rate) and GAE $\lambda$.
\begin{table}[ht] 
\caption{Hyperparameters for All Experiments} 
\centering
\resizebox{\linewidth}{!}{  
\begin{tabular}{cc|cc} 
\toprule
Method & Hyperparameter & Offline& Offline-to-Online\\
\hline
\multirow{8}{4em}{Filtered BC} & actor lr & 3e-3 & 3e-3 \\
& batch size & 128 & 128  \\
& rollout trajectories & - & 16 \\
& replay buffer size & - & 5000\\
& rollout temperature & - & 1.0 \\
& maximum gradient norm & 0.01 & 0.01 \\
& actor updates per iteration & 20 & 20 \\
& number of iterations for offline actor updates & 10 & 10 \\
\hline
\multirow{14}{4em}{\ouragentnospace} & actor lr & 3e-3 & 3e-3 \\
 & value function lr & 3e-3 & 3e-3 \\
  & instruction value function lr & 3e-3 & 3e-3 \\
  & instruction value function lr & 3e-3 & 3e-3 \\
& batch size & 128 & 128  \\
& rollout trajectories & - & 16 \\
& replay buffer size & - & 5000\\
& rollout temperature & - & 1.0 \\
& maximum gradient norm & 0.01 & 0.01 \\
& GAE $\lambda$ & 0.5 & 0.5 \\
& actor updates per iteration & 20 & 20 \\
& value function updates per iteration & 5 & 5 \\
& instruction value function updates per iteration & - & 5 \\
& number of iterations for offline actor updates & 10 & 10 \\
& number of iterations for offline value function updates & 20 & 20 \\
& number of iterations for offline instruction value function updates & - & 20 \\
\toprule
\end{tabular}}\label{tab:hyperparameters}
\caption{Hyperparameters for \ouragent and Filtered BC on both General and Web Shopping subset of AitW..}
\end{table}

\end{document}